\documentclass{article}

 \usepackage[preprint]{neurips_2026}

\usepackage[utf8]{inputenc} % allow utf-8 input
\usepackage[T1]{fontenc}    % use 8-bit T1 fonts
\usepackage{hyperref}       % hyperlinks
\usepackage{url}            % simple URL typesetting
\usepackage{booktabs}       % professional-quality tables
\usepackage{amsfonts}       % blackboard math symbols
\usepackage{nicefrac}       % compact symbols for 1/2, etc.
\usepackage{microtype}      % microtypography
\usepackage{xcolor}         % colors
\usepackage{amsmath}
\usepackage{amssymb}

\usepackage{booktabs}
\usepackage{tabularx}
\usepackage{array}

\usepackage{tikz}
\usepackage{adjustbox}
\usepackage{xcolor}
\usetikzlibrary{arrows.meta,positioning}
\usetikzlibrary{arrows.meta,positioning,fit}
\usetikzlibrary{calc,backgrounds}
\definecolor{deepblue}{RGB}{48,78,112}
\definecolor{teal}{RGB}{42,111,116}
\definecolor{purple}{RGB}{101,79,132}
\definecolor{orange}{RGB}{177,96,45}
\definecolor{green}{RGB}{62,124,86}
\definecolor{gold}{RGB}{166,116,32}
\definecolor{ink}{RGB}{42,46,52}

\definecolor{softblue}{RGB}{246,249,252}
\definecolor{softteal}{RGB}{245,250,249}
\definecolor{softpurple}{RGB}{249,247,252}
\definecolor{softorange}{RGB}{253,248,244}
\definecolor{softgreen}{RGB}{246,250,247}
\definecolor{softgold}{RGB}{253,250,242}
\definecolor{softgray}{RGB}{248,249,250}

\tikzset{inputbox/.style={
  draw=deepblue,
  fill=softblue,
  rounded corners=2.4pt,
  line width=0.90pt,
  align=center,
  font=\footnotesize,
  inner xsep=6pt,
  inner ysep=5pt,
  text=ink
}}

\tikzset{orchbox/.style={
  draw=deepblue,
  fill=softblue,
  rounded corners=2.4pt,
  line width=0.92pt,
  align=center,
  font=\footnotesize,
  inner xsep=5pt,
  inner ysep=4pt,
  text=ink
}}

\tikzset{agentbox/.style={
  draw=purple,
  fill=white,
  rounded corners=2.2pt,
  line width=0.72pt,
  align=center,
  font=\footnotesize,
  inner xsep=4pt,
  inner ysep=3.5pt,
  text=ink
}}

\tikzset{mcpbox/.style={
  draw=teal,
  fill=softteal,
  rounded corners=2.4pt,
  line width=0.88pt,
  align=center,
  font=\footnotesize,
  inner xsep=5pt,
  inner ysep=4pt,
  text=ink
}}

\tikzset{toolbox/.style={
  draw=teal,
  fill=white,
  rounded corners=2.2pt,
  line width=0.72pt,
  align=center,
  font=\footnotesize,
  inner xsep=4pt,
  inner ysep=3.3pt,
  text=ink
}}

\tikzset{policybox/.style={
  draw=green,
  fill=softgreen,
  rounded corners=2.2pt,
  line width=0.72pt,
  align=center,
  font=\footnotesize,
  inner xsep=4pt,
  inner ysep=3.3pt,
  text=ink
}}

\tikzset{sourcebox/.style={
  draw=gold,
  fill=softgold,
  rounded corners=2.4pt,
  line width=0.84pt,
  align=center,
  font=\footnotesize,
  inner xsep=5pt,
  inner ysep=4pt,
  text=ink
}}

\tikzset{sandboxbox/.style={
  draw=orange,
  fill=softorange,
  rounded corners=2.4pt,
  line width=0.84pt,
  align=center,
  font=\footnotesize,
  inner xsep=5pt,
  inner ysep=4pt,
  text=ink
}}

\tikzset{deliverybox/.style={
  draw=green,
  fill=softgreen,
  rounded corners=2.4pt,
  line width=0.84pt,
  align=center,
  font=\footnotesize,
  inner xsep=5pt,
  inner ysep=4pt,
  text=ink
}}

\tikzset{tieronebox/.style={
  draw=deepblue,
  fill=softblue,
  rounded corners=2.4pt,
  line width=0.82pt,
  align=center,
  font=\footnotesize,
  inner xsep=5pt,
  inner ysep=4pt,
  text=ink
}}

\tikzset{tiertwobox/.style={
  draw=teal,
  fill=softteal,
  rounded corners=2.4pt,
  line width=0.82pt,
  align=center,
  font=\footnotesize,
  inner xsep=5pt,
  inner ysep=4pt,
  text=ink
}}

\tikzset{evidencebox/.style={
  draw=green,
  fill=softgreen,
  rounded corners=2.4pt,
  line width=0.90pt,
  align=center,
  font=\footnotesize,
  inner xsep=5pt,
  inner ysep=4pt,
  text=ink
}}

\tikzset{platformframe/.style={
  draw=deepblue!75,
  fill=softpurple,
  rounded corners=3pt,
  line width=0.92pt
}}

\tikzset{generatedframe/.style={
  draw=teal!75,
  fill=softgray,
  rounded corners=3pt,
  line width=0.85pt
}}

\tikzset{crossbox/.style={
  draw=black!45,
  fill=softgray,
  rounded corners=2pt,
  line width=0.65pt,
  align=center,
  font=\footnotesize,
  inner xsep=5pt,
  inner ysep=3pt,
  text=ink
}}

\tikzset{flow/.style={
  ->,
  line width=0.86pt,
  draw=black!78
}}

\tikzset{sectionlabel/.style={
  font=\small\bfseries,
  anchor=west,
  text=ink
}}

\title{Graph, Loop, and Harness Engineering for Zero-Trust Agentic Data Engineering and Analytical Processing}

\author{%
  Sagar Srinivas Sakhinana, Venkataramana Runkana \\
  \texttt{sagar.sakhinana@tcs.com, venkat.runkana@tcs.com} \\
  Tata Research Development and Design Centre
}

\begin{document}

\maketitle

%%%%%%%%%%%%%%%%%%%%%%%%%%%%%%%%%%%%%%%%%%%%%%%%%%%%%%%%%%%%%%%%%%%%
%%%%%%%%%%%%%%%%%%%%%%%%%%%%%%%%%%%%%%%%%%%%%%%%%%%%%%%%%%%%%%%%%%%%
\begin{abstract}
Large language model agents increasingly automate data workflows, but end-to-end cloud data engineering and analytical execution require reliable coordination across code, data, infrastructure, and runtime environments. We present two zero-trust frameworks. \textbf{Zero-Trust Agentic Data Engineering} generates, deploys, and verifies complete cloud data-engineering solutions from natural-language tasks, with completion conditioned on repository, deployment, runtime, and policy evidence. \textbf{Zero-Trust Agentic OLAP} combines governed Data Preparation with verified Online Analytical Processing (OLAP), permitting production promotion only after validation and evidence-bound approval, and releasing analytical answers only after Same-Snapshot Execution, Exact Result Equivalence, deterministic grounding, and reflection. Both frameworks share three abstractions: \textbf{graph engineering} for evidence-gated workflow structure, \textbf{loop engineering} for bounded recovery, and \textbf{agent-harness engineering} for zero-trust execution. We evaluate both frameworks under nominal execution, controlled failures, bounded recovery, and policy-constrained conditions, measuring verified completion, recovery, authorization enforcement, production promotion, and verified OLAP execution.
\vspace{-1mm}
\end{abstract}
%%%%%%%%%%%%%%%%%%%%%%%%%%%%%%%%%%%%%%%%%%%%%%%%%%%%%%%%%%%%%%%%%%%%
%%%%%%%%%%%%%%%%%%%%%%%%%%%%%%%%%%%%%%%%%%%%%%%%%%%%%%%%%%%%%%%%%%%%

%%%%%%%%%%%%%%%%%%%%%%%%%%%%%%%%%%%%%%%%%%%%%%%%%%%%%%%%%%%%%%%%%%%%
%%%%%%%%%%%%%%%%%%%%%%%%%%%%%%%%%%%%%%%%%%%%%%%%%%%%%%%%%%%%%%%%%%%%
\section{Introduction}
\label{sec:introduction}
\vspace{-1mm}
Recent software-engineering (SWE) agents extend automation from code generation to tool-using repository interaction, execution, repair, and validation~\cite{yao2023react,wu2024autogen,jimenez2024swebench,yang2024sweagent,zhang2024autocoderover,wang2024codeact,wang2025openhands,xia2025agentless}. In parallel, data-engineering agents increasingly automate data preparation, pipeline generation, execution, and governance~\cite{zhang2023datacopilot,fan2024autoprep,jin2025eltbench,siva2026kraig,he2026dataflow,kirubakaran2025governing}. Cloud-native data engineering spans batch extract-transform-load (ETL) and extract-load-transform (ELT) pipelines for reporting and reconciliation; real-time streaming pipelines for e-commerce, payments, and mobility; Change Data Capture (CDC) pipelines for operational-data replication and synchronization; data lakes, lakehouses, data warehouses, and data marts for analytics and artificial intelligence (AI); Internet of Things (IoT) telemetry pipelines for operational monitoring; and real-time analytics and machine learning (ML) pipelines for fraud detection, recommendations, forecasting, and predictive maintenance. Agentic AI frameworks can automate these workloads through code generation, infrastructure provisioning, pipeline execution, and analytical querying. However, failures can arise across software, infrastructure, orchestration, runtime, data, and analytics because of configuration errors, Identity and Access Management (IAM; controls identities, roles, and permissions) misconfigurations, non-idempotent workflows, schema and data-quality violations, resource and scaling constraints, insufficient observability, or semantically incorrect analytical results. These failures can disrupt services, violate security policies, increase cloud costs, or produce invalid data and analytical outputs. Reliable operation therefore requires verified Data Engineering and trustworthy analytical execution. Data Engineering covers batch and streaming ingestion, schema evolution, storage, transformation, orchestration, pipeline execution, schema and data-quality validation, policy enforcement, lineage capture, authorization, and controlled production promotion. Online Analytical Processing (OLAP) operates over governed data products in cloud data warehouses, supporting multidimensional analytical SQL for filtering, aggregation, grouping, ranking, trend analysis, and verification of computed results and answers. Cloud operation additionally requires secrets management, network isolation, artifact integrity, Policy as Code (PaC), elastic scaling, monitoring, recovery, and rollback. Trustworthy agentic automation must therefore operate under isolated execution, policy enforcement, evidence-gated progression, and bounded recovery. This motivates transforming natural-language Data Engineering and OLAP tasks---for example, `Build and deploy a real-time e-commerce platform that ingests clickstream, order, and payment events, processes them through scalable pipelines, exposes backend APIs, and provides a frontend dashboard for sales and operational metrics,'' or `Perform OLAP over governed data to support multidimensional analytical queries across domains; for example, in e-commerce, analyze revenue by region, conversion rates, top-selling products, customer segments, and failed-payment trends''---into validated code repositories implementing Continuous Integration, Continuous Delivery, and Continuous Testing (CI/CD/CT), cloud infrastructure, data pipelines, governed data products, and verified analytics. To address this problem, we propose two complementary frameworks. \textbf{Zero-Trust Agentic Data Engineering} transforms natural-language tasks into complete cloud data-engineering solutions spanning application code, data pipelines, orchestration, infrastructure, deployment, and runtime operation. \textbf{Zero-Trust Agentic OLAP} transforms natural-language analytical tasks over governed data products into verified analytical answers through policy-constrained SQL generation and execution, multidimensional analysis, and verification of computed results and answers. Both frameworks are organized around three engineering abstractions: \textbf{graph engineering}, \textbf{loop engineering}, and \textbf{agent-harness engineering}. Graph engineering represents long-horizon workflows as explicit stages, dependencies, transitions, verification predicates, recovery paths, and terminal conditions, ensuring that progression occurs only when the required evidence is established. Loop engineering governs bounded diagnosis, repair, re-planning, and retry when verification conditions fail, preventing unbounded execution, privilege expansion, or resource consumption. Agent-harness engineering establishes a zero-trust boundary around agent actions through workload identity, authorization, policy enforcement, mediated tool access, isolated execution, tenant and data isolation, execution bounds, and evidence capture. Together, these abstractions isolate consequential execution, condition workflow progression on independently established evidence, and confine repairable failures to bounded recovery paths. Zero-Trust Agentic Data Engineering reaches verified completion only when repository validity, cloud deployment, runtime behavior, and policy compliance are supported by retained evidence. Zero-Trust Agentic OLAP permits production promotion only after Preparation Validation and evidence-bound approval, and releases a Verified Answer only after Same-Snapshot Execution, Exact Result Equivalence, deterministic grounding, and reflection. Across both frameworks, terminal success is therefore determined by independently established verification evidence rather than agent-reported success.
\vspace{-2mm}
\begin{itemize}
\item We introduce two distinct frameworks: \textbf{Zero-Trust Agentic Data Engineering} for generating, deploying, and verifying complete cloud data-engineering solutions, and \textbf{Zero-Trust Agentic OLAP} for governed Data Preparation and verified Online Analytical Processing over Governed Data Products.
\vspace{-1mm}
\item We formulate \textbf{graph engineering}, \textbf{loop engineering}, and \textbf{agent-harness engineering} as shared abstractions for structuring long-horizon agent workflows, bounded recovery, and zero-trust execution, with workflow progression conditioned on explicit verification evidence.
\vspace{-5mm}
\item We construct framework-specific benchmark suites and empirically evaluate both frameworks under nominal execution, controlled failures, bounded recovery, and policy-constrained conditions, examining verified completion, recovery, authorization enforcement, production promotion, and verified OLAP execution.
\end{itemize}

\vspace{-2mm}
The subsequent sections detail the frameworks, experimental setup, benchmark suites, and empirical evaluation.
\vspace{-2mm}
%%%%%%%%%%%%%%%%%%%%%%%%%%%%%%%%%%%%%%%%%%%%%%%%%%%%%%%%%%%%%%%%%%%%
%%%%%%%%%%%%%%%%%%%%%%%%%%%%%%%%%%%%%%%%%%%%%%%%%%%%%%%%%%%%%%%%%%%%

%%%%%%%%%%%%%%%%%%%%%%%%%%%%%%%%%%%%%%%%%%%%%%%%%%%%%%%%%%%%%%%%%%%%
%%%%%%%%%%%%%%%%%%%%%%%%%%%%%%%%%%%%%%%%%%%%%%%%%%%%%%%%%%%%%%%%%%%%
\section{Overall Framework}
\label{sec:overall-framework}
\vspace{-2mm}
Zero-Trust Agentic Data Engineering generates, deploys, and verifies complete cloud data-engineering solutions from natural-language tasks, spanning applications, services, pipelines,  infrastructure, IaC, and tests. Zero-Trust Agentic OLAP focuses on governed data preparation and verified analytical querying through sandboxed transformation, evidence-gated promotion, controlled OLAP execution, and grounded answer generation.

\clearpage
\newpage

%%%%%%%%%%%%%%%%%%%%%%%%%%%%%%%%%%%%%%%%%%%%%%%%%%%%%%%%%%%%%%%%%%%%
%%%%%%%%%%%%%%%%%%%%%%%%%%%%%%%%%%%%%%%%%%%%%%%%%%%%%%%%%%%%%%%%%%%%
\begin{figure*}[ht!]
\centering
\begin{adjustbox}{
  max width=\textwidth,
  max totalheight=0.92\textheight,
  keepaspectratio
}

\begin{tikzpicture}[x=1cm,y=1cm]

% ------------------------------------------------------------------
% NATURAL-LANGUAGE TASK
% ------------------------------------------------------------------

\node[
  sectionlabel
] at (0.25,1.12)
{
  Natural-Language Task
};

\draw[
  deepblue!35,
  line width=0.75pt
]
(0.25,0.92)
--
(14.45,0.92);

\node[
  inputbox,
  minimum width=84mm,
  text width=80mm,
  minimum height=10mm
] (task) at (7.20,0.00)
{
  \textbf{User Data-Engineering Task}
};

% ------------------------------------------------------------------
% AGENTIC CONTROL PLANE
% ------------------------------------------------------------------

\node[
  sectionlabel
] at (-0.20,-1.28)
{
  Agentic Control Plane -- Regular Agent-Hosting GKE
};

\draw[
  deepblue!35,
  line width=0.75pt
]
(0.25,-1.52)
--
(14.45,-1.52);

\node[
  orchbox,
  minimum width=104mm,
  text width=100mm,
  minimum height=9mm
] (orch) at (7.20,-2.45)
{
  \textbf{Graph Orchestrator (Google ADK)}
  $\cdot$
  A2A delegation
  $\cdot$
  Evidence-Gated Workflow
};

\node[
  agentbox,
  minimum width=30mm,
  text width=26mm,
  minimum height=11mm
] (plan) at (1.90,-3.85)
{
  \textbf{Planning Agent}\\
  architecture / plan
};

\node[
  agentbox,
  minimum width=30mm,
  text width=26mm,
  minimum height=11mm
] (gen) at (5.25,-3.85)
{
  \textbf{Generation Agent}\\
  full-project code
};

\node[
  agentbox,
  minimum width=30mm,
  text width=26mm,
  minimum height=11mm
] (review) at (8.60,-3.85)
{
  \textbf{Review Agent}\\
  code / config review
};

\node[
  agentbox,
  minimum width=30mm,
  text width=26mm,
  minimum height=11mm
] (execagent) at (11.95,-3.85)
{
  \textbf{Execution / Deployment Agent}\\
  sandbox / cloud actions
};

\node[
  agentbox,
  minimum width=30mm,
  text width=26mm,
  minimum height=11mm
] (verifyagent) at (1.90,-5.35)
{
  \textbf{Verification Agent}\\
  verification gates
};

\node[
  agentbox,
  minimum width=30mm,
  text width=26mm,
  minimum height=11mm
] (monitor) at (5.25,-5.35)
{
  \textbf{Runtime Monitoring Agent}\\
  runtime evidence
};

\node[
  agentbox,
  minimum width=30mm,
  text width=26mm,
  minimum height=11mm
] (reflect) at (8.60,-5.35)
{
  \textbf{Reflection / Diagnosis Agent}\\
  failure diagnosis
};

\node[
  agentbox,
  minimum width=30mm,
  text width=26mm,
  minimum height=11mm
] (repair) at (11.95,-5.35)
{
  \textbf{Repair / Re-plan Agent}\\
  bounded repair / retry
};

% Only the orchestrator and agent workloads are enclosed by
% the regular agent-hosting GKE boundary.
\begin{scope}[on background layer]

\node[
  platformframe,
  fit=
  (orch)
  (plan)
  (gen)
  (review)
  (execagent)
  (verifyagent)
  (monitor)
  (reflect)
  (repair),
  inner xsep=8pt,
  inner ysep=8pt
] (platform) {};

\end{scope}

\draw[flow]
(task.south)
--
(task.south |- platform.north);

% ------------------------------------------------------------------
% CONTROLLED TOOL ACCESS THROUGH MCP
% ------------------------------------------------------------------

\node[
  sectionlabel
] at (0.25,-6.65)
{
  Controlled Tool Access through MCP
};

\draw[
  deepblue!35,
  line width=0.75pt
]
(0.25,-6.90)
--
(14.45,-6.90);

\node[
  mcpbox,
  minimum width=118mm,
  text width=114mm,
  minimum height=10mm
] (mcp) at (7.20,-7.95)
{
  \textbf{Tenant-Scoped MCP Layer}
  $\cdot$
  controlled agent-to-tool access
  $\cdot$
  policy-scoped operations
};

% Explicitly align the incoming MCP arrow with the center
% of the MCP box so that it remains perfectly vertical.
\coordinate
(mcpin)
at
(mcp.north |- platform.south);

\draw[flow]
(mcpin)
--
(mcp.north);

% Controlled tool environments
\node[
  toolbox,
  minimum width=32mm,
  text width=28mm,
  minimum height=10mm
] (vscode) at (1.70,-9.35)
{
  \textbf{VS Code Sandbox}\\
  inspect / edit / test
};

\node[
  toolbox,
  minimum width=32mm,
  text width=28mm,
  minimum height=10mm
] (chrome) at (5.35,-9.35)
{
  \textbf{Chrome Sandbox}\\
  research / browser checks
};

\node[
  toolbox,
  minimum width=32mm,
  text width=28mm,
  minimum height=10mm
] (cloudtools) at (9.00,-9.35)
{
  \textbf{Cloud / Execution Tools}\\
  build / deploy / verify
};

\node[
  policybox,
  minimum width=32mm,
  text width=28mm,
  minimum height=10mm
] (harness) at (12.65,-9.35)
{
  \textbf{Zero-Trust Harness}\\
  auth / policy / tenant scope
};

% Separate MCP-mediated tool-access boundary.
\begin{scope}[on background layer]

\node[
  platformframe,
  fit=
  (mcp)
  (vscode)
  (chrome)
  (cloudtools)
  (harness),
  inner xsep=8pt,
  inner ysep=8pt
] (toolaccess) {};

\end{scope}

% ------------------------------------------------------------------
% GENERATED FULL-PROJECT SOURCE
% ------------------------------------------------------------------

\node[
  sectionlabel
] at (0.25,-10.75)
{
  Generated Full-Project Source
};

\draw[
  gold!40,
  line width=0.75pt
]
(0.25,-11.00)
--
(14.45,-11.00);

\node[
  sourcebox,
  minimum width=118mm,
  text width=114mm,
  minimum height=11mm
] (source) at (7.20,-11.95)
{
  \textbf{Complete Generated Project}\\
  Frontend
  $\cdot$
  Backend
  $\cdot$
  Data Pipelines
  $\cdot$
  IaC
  $\cdot$
  Tests
};

% Keeps the outgoing arrow from the tool-access layer vertical.
\coordinate
(toolaccessout)
at
(source.north |- toolaccess.south);

\draw[flow]
(toolaccessout)
--
(source.north);

% ------------------------------------------------------------------
% ISOLATED EXECUTION AND TRUSTED DELIVERY
% ------------------------------------------------------------------

\node[
  sectionlabel
] at (0.25,-13.25)
{
  Isolated Execution and Trusted Delivery
};

\draw[
  deepblue!35,
  line width=0.75pt
]
(0.25,-13.50)
--
(14.45,-13.50);

\node[
  sandboxbox,
  minimum width=57mm,
  text width=53mm,
  minimum height=13mm
] (gvisor) at (3.65,-14.65)
{
  \textbf{Per-Run GKE Sandbox with gVisor}\\
  isolated execution
  $\cdot$
  tests
  $\cdot$
  IaC validation
};

\node[
  deliverybox,
  minimum width=57mm,
  text width=53mm,
  minimum height=13mm
] (delivery) at (10.45,-14.65)
{
  \textbf{Trusted Build and Delivery}\\
  build / test
  $\cdot$
  security / IaC
  $\cdot$
  provenance
  $\cdot$
  deploy
};

\coordinate
(sourceout)
at
(gvisor.north |- source.south);

\draw[flow]
(sourceout)
--
(gvisor.north);

\draw[flow]
(gvisor.east)
--
(delivery.west);

% ------------------------------------------------------------------
% GENERATED TARGET CLOUD SOLUTION
% ------------------------------------------------------------------

\node[
  sectionlabel
] at (0.25,-16.10)
{
  Generated Target Two-Tier Cloud Solution
};

\draw[
  teal!45,
  line width=0.75pt
]
(0.25,-16.35)
--
(14.45,-16.35);

\node[
  tieronebox,
  minimum width=63mm,
  text width=59mm,
  minimum height=14mm
] (tier1) at (3.60,-17.55)
{
  \textbf{Tier 1 -- Edge + Web Frontend}\\
  DNS / HTTPS / WAF
  $\cdot$
  Frontend
  $\cdot$
  CDN
};

\node[
  tiertwobox,
  minimum width=63mm,
  text width=59mm,
  minimum height=14mm
] (tier2) at (10.55,-17.55)
{
  \textbf{Tier 2 -- Backend + Data Engineering}\\
  APIs / Services
  $\cdot$
  Pipelines
  $\cdot$
  Storage
  $\cdot$
  IaC
};

\draw[flow]
(tier1.east)
--
(tier2.west);

\begin{scope}[on background layer]

\node[
  generatedframe,
  fit=(tier1)(tier2),
  inner xsep=8pt,
  inner ysep=8pt
] (generated) {};

\end{scope}

\draw[flow]
(delivery.south)
--
(delivery.south |- generated.north);

% ------------------------------------------------------------------
% EVIDENCE-GATED VERIFICATION
% ------------------------------------------------------------------

\node[
  evidencebox,
  minimum width=128mm,
  text width=124mm,
  minimum height=14mm
] (evidence) at (7.10,-19.65)
{
  \textbf{Cloud Deployment and Runtime Verification}\\
  $\mathrm{RepoValid}
  \land
  \mathrm{DeployVerified}
  \land
  \mathrm{RuntimeVerified}
  \land
  \mathrm{PolicyOK}$\\
  \textbf{PASS:}
  completion / promotion
  \qquad
  \textbf{FAIL:}
  bounded recovery
};

\coordinate
(generatedout)
at
(evidence.north |- generated.south);

\draw[flow]
(generatedout)
--
(evidence.north);

% ------------------------------------------------------------------
% CROSS-CUTTING CONTROLS
% ------------------------------------------------------------------

\node[
  crossbox,
  minimum width=128mm,
  text width=124mm,
  minimum height=8mm
] (cross) at (7.10,-21.05)
{
  \textbf{Cross-Cutting:}
  IAM / WIF
  $\cdot$
  RBAC / ABAC
  $\cdot$
  Secrets
  $\cdot$
  Policy
  $\cdot$
  Network / Runtime Security
  $\cdot$
  Observability / Audit
};

\end{tikzpicture}

\end{adjustbox}
\vspace{-1mm}
\caption{
\textbf{Overview of the proposed agentic data-engineering framework.}
ADK orchestrates agents on GKE, A2A supports delegation, and MCP controls tool access. Generated workloads run in a per-run gVisor sandbox with bounded recovery.
}
\label{fig:agentic-two-tier-data-engineering}
\vspace{-2mm}
\end{figure*}
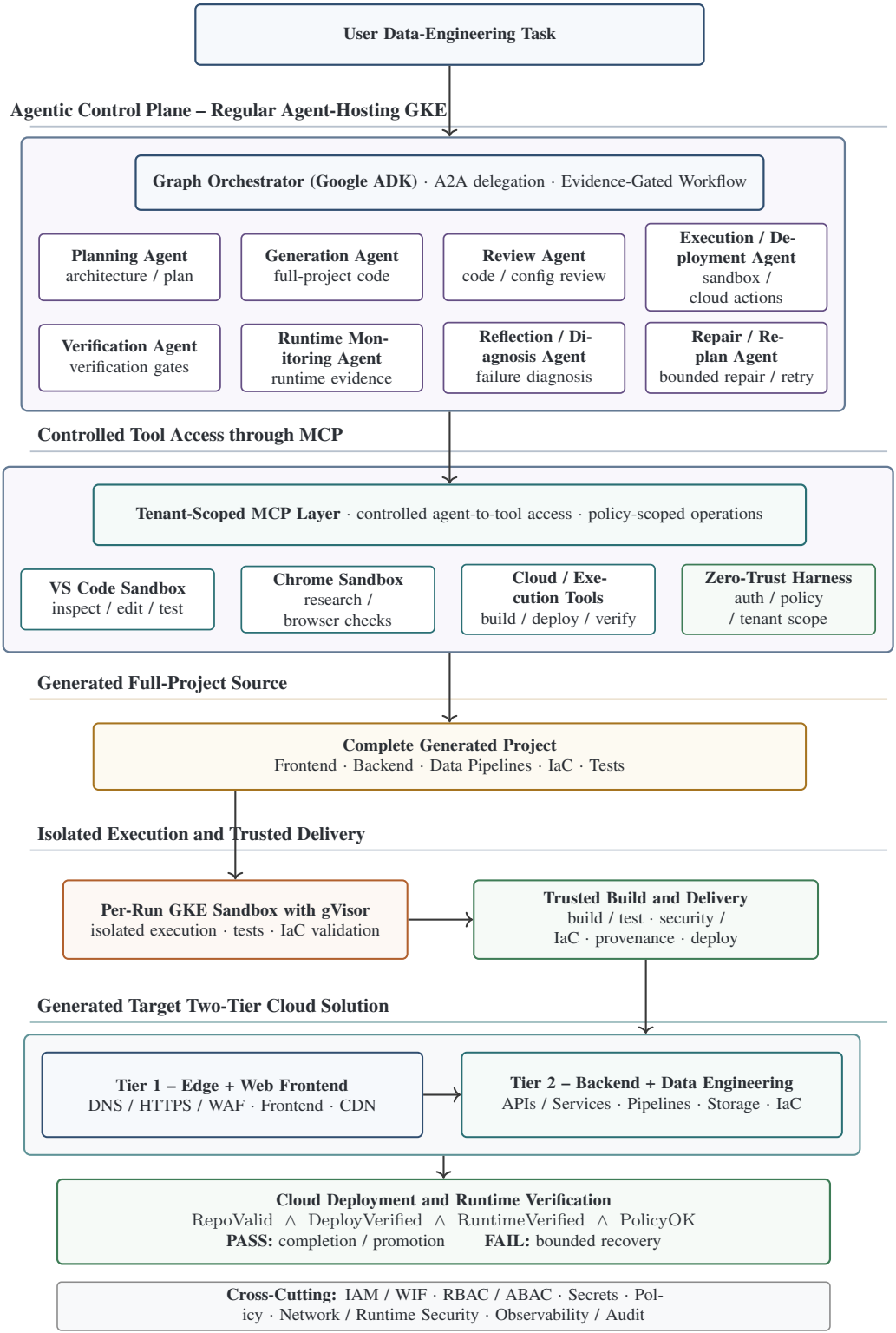

\subsection{Zero-Trust Agentic Data Engineering}
\vspace{-2mm}
The proposed Zero-Trust Agentic Data Engineering framework generates and verifies complete, deployable data-engineering solutions from natural-language tasks. For each task, it synthesizes the frontend, backend services and Application Programming Interfaces (APIs), data pipelines and orchestration, cloud infrastructure, Infrastructure as Code (IaC), and tests, and validates the resulting system through isolated execution, trusted delivery, cloud deployment verification, runtime verification, and policy enforcement. Figure~\ref{fig:agentic-two-tier-data-engineering} provides a system-level overview of the framework, spanning natural-language task ingestion, graph-orchestrated multi-agent generation, controlled tool access, isolated execution, trusted delivery, deployment of the generated two-tier cloud solution, and evidence-gated completion. The framework is organized around three complementary abstractions: graph engineering, loop engineering, and agent-harness engineering. Graph engineering structures the long-horizon workflow as explicit stages, dependencies, transitions, verification gates, recovery paths, and terminal conditions. Loop engineering governs bounded diagnosis, repair, re-planning, and retry. Agent-harness engineering constrains agent actions through identity, authorization, policy enforcement, controlled tool access, isolation, and evidence collection. (a) \textbf{Graph-Orchestrated Multi-Agent Control.} A natural-language task enters a shared control plane running on the regular agent-hosting Google Kubernetes Engine (GKE) layer. A Graph Orchestrator implemented using the Google Agent Development Kit (ADK) acts as the parent agent, maintains workflow state, coordinates specialized agents, and governs evidence-gated progression. Agent2Agent (A2A) communication supports structured delegation and information exchange among agents. (i) The Planning Agent derives the system architecture and implementation plan. (ii) The Generation Agent produces the complete project. (iii) The Review Agent evaluates generated code, configuration, dependencies, and infrastructure definitions. (iv) The Execution and Deployment Agent coordinates isolated execution and the trusted delivery path. (v) The Verification Agent evaluates repository, cloud deployment, and runtime verification conditions. (vi) The Runtime Monitoring Agent collects runtime evidence, including application health, frontend-to-backend communication, API responses, data-pipeline execution, integration and end-to-end test results, logs, metrics, traces, and policy observations. (vii) The Reflection/Diagnosis and Repair/Re-plan Agents analyze verification failures, generate corrective changes, revise the implementation plan, and initiate bounded retries. Recovery is bounded by limits on attempts, time, tokens, tool calls, cost, privilege, and blast radius, with explicit termination when the recovery budget is exhausted. (b) \textbf{Controlled Tool Access and Isolated Execution.} All agents execute on the regular agent-hosting GKE layer and access tool and execution capabilities through a tenant-scoped Model Context Protocol (MCP) layer. MCP provides a standardized agent-to-tool interface, while the surrounding zero-trust controls determine which tools, resources, and operations may be invoked. The MCP layer exposes controlled access to (i) a Visual Studio Code (VS Code) Sandbox for repository inspection, code modification, and testing; (ii) a Chrome Sandbox for controlled web research and browser-based validation; and (iii) a separate per-run GKE Sandbox with gVisor for isolated execution, testing, and Infrastructure as Code (IaC) validation of generated workloads. In this execution model, (i) ADK provides the agent runtime and graph orchestration, (ii) A2A supports coordination among the parent and specialized agents, (iii) MCP provides controlled agent-to-tool interaction, and GKE Sandbox with gVisor provides the isolated runtime boundary for generated workloads. (c) \textbf{Trusted Delivery and Generated Cloud Deployment.}
Candidate code repository revisions that satisfy sandbox checks enter the trusted delivery path coordinated by the Execution and Deployment Agent. There, the immutable source revision is independently rebuilt, subjected to build, security, and IaC validation, associated with a Software Bill of Materials (SBOM) and build provenance, published to Artifact Registry, admitted through Binary Authorization, and deployed through Cloud Deploy. The generated solution is deployed as a two-tier cloud architecture. (i) Tier 1 realizes the generated frontend through Domain Name System (DNS), Hypertext Transfer Protocol Secure (HTTPS), Transport Layer Security (TLS), Web Application Firewall (WAF), load balancing, hosting, and Content Delivery Network (CDN) configuration, and exposes task-specific dashboards, data visualizations, data views, and query and result exploration. (ii) Tier 2 integrates backend services and APIs with data ingestion, transformation, orchestration, storage, processing, analytics, and data-serving components. (d) \textbf{Zero-Trust Enforcement and Evidence-Gated Completion.} Consequential cloud operations are governed by a zero-trust agent harness using Identity and Access Management (IAM), Workload Identity Federation (WIF), Role-Based Access Control (RBAC), and Attribute-Based Access Control (ABAC). IAM defines workload identities and permissions over cloud resources. WIF enables workloads to obtain short-lived federated credentials without long-lived service-account keys. RBAC constrains access according to assigned roles. ABAC evaluates authorization using contextual attributes such as tenant, workload identity, target resource, requested operation, and execution context. Open Policy Agent (OPA) evaluates declarative authorization and governance policies, while Gatekeeper enforces admission policies for Kubernetes resources. Secret Manager protects credentials, tokens, and sensitive configuration. Cilium and Kubernetes NetworkPolicy constrain network communication among workloads and services, while Falco detects anomalous or policy-violating runtime behavior. Tenant-scoped state and row-level security preserve data isolation, while OpenTelemetry, Cloud Logging, Cloud Monitoring, metrics, traces, and audit records provide observability and retain execution and verification evidence. Workflow completion or promotion is permitted only when repository validity, deployment verification, runtime verification, and policy compliance are supported by retained evidence; otherwise, the workflow enters bounded diagnosis, repair, and re-planning until the verification conditions are satisfied or the recovery budget is exhausted. Overall, the framework progresses from natural-language task interpretation through graph-orchestrated multi-agent generation, MCP-mediated tool access, isolated execution in GKE Sandbox with gVisor, trusted build and delivery, deployment of the generated two-tier cloud solution, and independent deployment and runtime verification, terminating in either verified completion or bounded recovery.

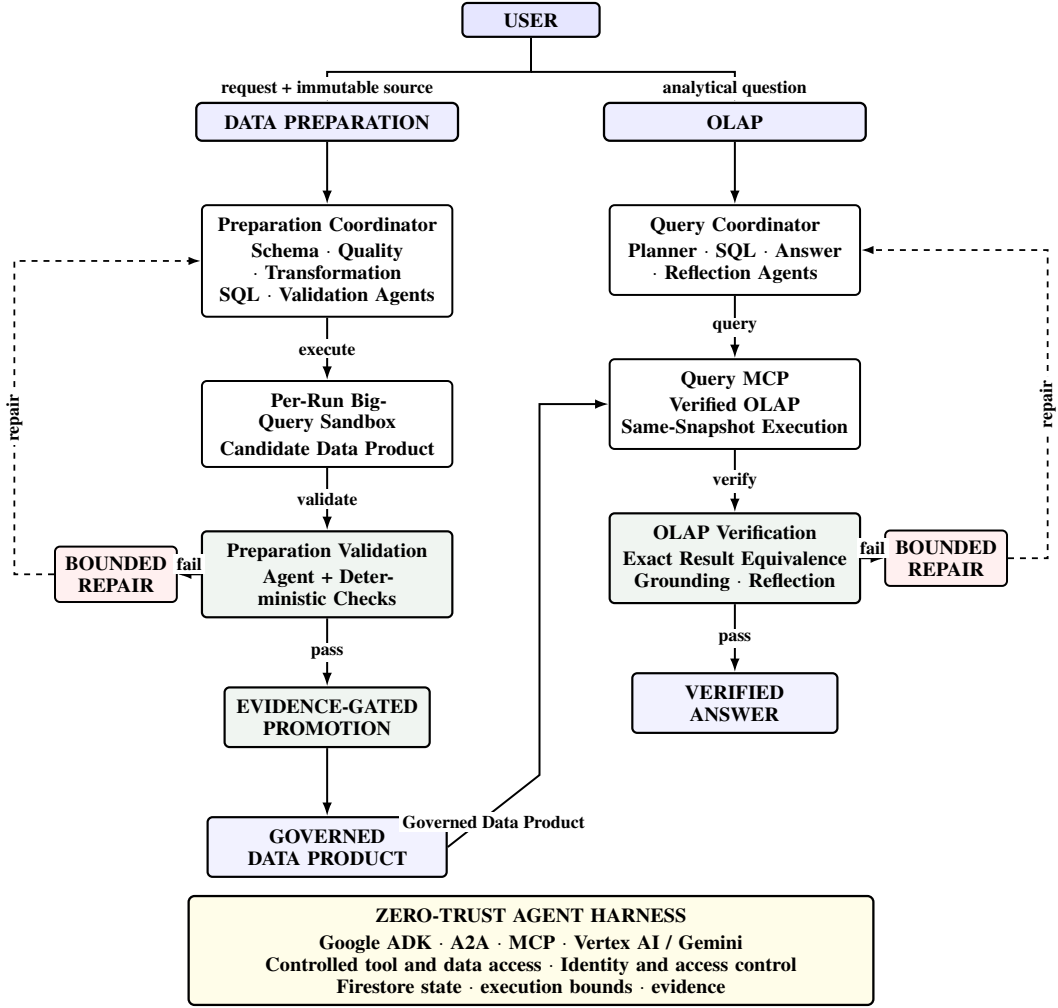
\begin{figure*}[ht!]
\vspace{-2mm}
\centering
\begin{adjustbox}{max width=\textwidth,keepaspectratio}
\begin{tikzpicture}[
  >=Latex,
  flow/.style={-{Latex[length=2.4mm,width=1.7mm]},line width=1.0pt},
  feedback/.style={-{Latex[length=2.2mm,width=1.5mm]},dashed,line width=0.9pt},
  edgeLabel/.style={fill=white,inner sep=1.7pt,font=\small\bfseries},
  userbox/.style={draw,rounded corners=2.5pt,line width=1.1pt,fill=blue!10,
                 align=center,inner xsep=6pt,inner ysep=5pt,
                 font=\normalsize\bfseries},
  lanehead/.style={draw,rounded corners=3pt,line width=1.15pt,fill=blue!7,
                  align=center,inner xsep=6pt,inner ysep=6pt,
                  font=\normalsize\bfseries},
  stagebox/.style={draw,rounded corners=2.5pt,line width=1.0pt,fill=white,
                  align=center,inner xsep=6pt,inner ysep=6pt,
                  font=\normalsize\bfseries},
  verifybox/.style={draw,rounded corners=2.5pt,line width=1.1pt,fill=green!8,
                   align=center,inner xsep=6pt,inner ysep=6pt,
                   font=\normalsize\bfseries},
  failbox/.style={draw,rounded corners=2.5pt,line width=1.0pt,fill=red!6,
                 align=center,inner xsep=5pt,inner ysep=5pt,
                 font=\normalsize\bfseries},
  resultbox/.style={draw,rounded corners=2.5pt,line width=1.1pt,fill=blue!5,
                   align=center,inner xsep=6pt,inner ysep=6pt,
                   font=\normalsize\bfseries},
  harness/.style={draw,rounded corners=3pt,line width=1.15pt,fill=yellow!10,
                 align=center,inner xsep=7pt,inner ysep=6pt,
                 font=\normalsize\bfseries}
]

% USER AND WORKFLOW HEADINGS
\node[userbox,text width=20mm] (user) {USER};

\node[lanehead,text width=42mm]
  (prepH) at ($(user)+(-36mm,-18mm)$)
  {DATA PREPARATION};

\node[lanehead,text width=42mm]
  (anaH) at ($(user)+(36mm,-18mm)$)
  {OLAP};

\coordinate (split) at ($(user.south)+(0,-6mm)$);

\draw[flow]
  (user.south) -- (split)
  -| node[edgeLabel,pos=0.78]{request + immutable source}
  (prepH.north);

\draw[flow]
  (split)
  -| node[edgeLabel,pos=0.78]{analytical question}
  (anaH.north);

% DATA PREPARATION
\node[stagebox,text width=40mm,below=11mm of prepH] (prepAgents) {
Preparation Coordinator\\[0.8mm]
Schema $\cdot$ Quality $\cdot$ Transformation\\
SQL $\cdot$ Validation Agents};

\node[stagebox,text width=40mm,below=11mm of prepAgents] (sandbox) {
Per-Run BigQuery Sandbox\\[0.8mm]
Candidate Data Product};

\node[verifybox,text width=40mm,below=11mm of sandbox] (pval) {
Preparation Validation\\[0.8mm]
Agent + Deterministic Checks};

\node[failbox,text width=18mm] (prepair)
  at ($(pval.west)+(-15mm,0)$) {
BOUNDED\\
REPAIR};

\node[verifybox,text width=32mm,below=12mm of pval] (promotion) {
EVIDENCE-GATED\\
PROMOTION};

\node[resultbox,text width=38mm,below=12mm of promotion] (product) {
GOVERNED\\
DATA PRODUCT};

\draw[flow]
  (prepH) -- (prepAgents);

\draw[flow]
  (prepAgents) --
  node[edgeLabel]{execute}
  (sandbox);

\draw[flow]
  (sandbox) --
  node[edgeLabel]{validate}
  (pval);

\draw[flow]
  (pval.west) --
  node[edgeLabel,above]{fail}
  (prepair.east);

\draw[flow]
  (pval) --
  node[edgeLabel]{pass}
  (promotion);

\draw[flow]
  (promotion) -- (product);

% Preparation repair path
\coordinate (pleft) at ($(prepair.west)+(-7mm,0)$);

\draw[feedback]
  (prepair.west)
  -- (pleft)
  --
  node[edgeLabel,rotate=90]{repair}
  (pleft |- prepAgents.west)
  -- (prepAgents.west);

% OLAP
\node[stagebox,text width=40mm,below=11mm of anaH] (anaAgents) {
Query Coordinator\\[0.8mm]
Planner $\cdot$ SQL $\cdot$ Answer\\
$\cdot$ Reflection Agents};

\node[stagebox,text width=40mm,below=11mm of anaAgents] (queryExec) {
Query MCP\\[0.8mm]
Verified OLAP\\
Same-Snapshot Execution};

\node[verifybox,text width=40mm,below=11mm of queryExec] (qval) {
OLAP Verification\\[0.8mm]
Exact Result Equivalence\\
Grounding $\cdot$ Reflection};

\node[resultbox,text width=32mm,below=12mm of qval] (verified) {
VERIFIED\\
ANSWER};

\node[failbox,text width=18mm] (qrepair)
  at ($(qval.east)+(15mm,0)$) {
BOUNDED\\
REPAIR};

\draw[flow]
  (anaH) -- (anaAgents);

\draw[flow]
  (anaAgents) --
  node[edgeLabel]{query}
  (queryExec);

\draw[flow]
  (queryExec) --
  node[edgeLabel]{verify}
  (qval);

\draw[flow]
  (qval.east) --
  node[edgeLabel,above]{fail}
  (qrepair.west);

\draw[flow]
  (qval) --
  node[edgeLabel]{pass}
  (verified);

% OLAP repair path
\coordinate (aright) at ($(qrepair.east)+(7mm,0)$);

\draw[feedback]
  (qrepair.east)
  -- (aright)
  --
  node[edgeLabel,rotate=90]{repair}
  (aright |- anaAgents.east)
  -- (anaAgents.east);

% GOVERNED DATA PRODUCT TO OLAP
\coordinate (dataTurn)
  at ($(product.east)!0.5!(verified.west)$);

\draw[flow]
  (product.east)
  --
  node[edgeLabel,below]{Governed Data Product}
  (dataTurn)
  |- (queryExec.west);

% SHARED ZERO-TRUST AGENT HARNESS
\node[harness,text width=116mm,anchor=north] (harness)
  at ($(product.south)+(36mm,-3mm)$) {
ZERO-TRUST AGENT HARNESS\\[1mm]
Google ADK $\cdot$ A2A $\cdot$ MCP $\cdot$ Vertex AI / Gemini\\
Controlled tool and data access $\cdot$ Identity and access control\\
Firestore state $\cdot$ execution bounds $\cdot$ evidence};

\end{tikzpicture}
\end{adjustbox}

\caption{\textbf{Overall framework for Zero-Trust Agentic OLAP.} The framework comprises two evidence-gated multi-agent workflows: \textbf{Data Preparation}, which transforms an immutable source dataset into a validated Governed Data Product through sandboxed execution, Preparation Validation, and Evidence-Gated Promotion; and \textbf{OLAP}, which answers natural-language analytical questions over that product through verified Online Analytical Processing. Repairable failures use bounded stage-local recovery, while the shared Zero-Trust Agent Harness provides orchestration, delegation, controlled tool and data access, identity and access control, workflow state, execution bounds, and evidence retention.
}
\label{fig:overall-agentic-framework}
\vspace{-1mm}
\end{figure*}

\subsection{Zero-Trust Agentic OLAP}
\begin{figure*}[ht!]
\centering
\begin{adjustbox}{max width=\textwidth,keepaspectratio}
\begin{tikzpicture}[
  >=Latex,
  flow/.style={-{Latex[length=3.2mm,width=2.25mm]},line width=1.15pt,
               shorten >=1.2pt,shorten <=1pt},
  edgeLabel/.style={fill=white,inner sep=1.9pt,font=\normalsize\bfseries},
  title/.style={draw,rounded corners=3pt,line width=1.1pt,fill=blue!8,
               align=center,inner xsep=8pt,inner ysep=6pt,
               font=\normalsize\bfseries},
  box/.style={draw,rounded corners=2.5pt,line width=1.0pt,fill=white,
             align=center,inner xsep=7pt,inner ysep=6pt,
             font=\normalsize\bfseries},
  project/.style={draw,rounded corners=4pt,line width=1.15pt,inner sep=8pt},
  gate/.style={draw,rounded corners=2.5pt,line width=1.1pt,fill=green!8,
              align=center,inner xsep=7pt,inner ysep=6pt,
              font=\normalsize\bfseries},
  boundary/.style={align=center,font=\small\bfseries,
                  fill=white,inner sep=2pt},
  identity/.style={draw,rounded corners=2.5pt,line width=1.0pt,fill=gray!7,
                  align=center,inner xsep=7pt,inner ysep=5pt,
                  font=\normalsize\bfseries},
  zt/.style={draw,rounded corners=3pt,line width=1.1pt,fill=yellow!10,
            align=center,inner xsep=8pt,inner ysep=7pt,
            font=\normalsize\bfseries}
]

% ================================================================
% REQUESTS
% ================================================================
\node[title,text width=35mm] (preq)
  at (-46mm,0) {
PREPARATION\\
REQUEST};

\node[title,text width=35mm] (qreq)
  at (46mm,0) {
ANALYTICS\\
QUESTION};

% ================================================================
% CONTROL PROJECT
% ================================================================
\node[title,text width=72mm] (controlTitle)
  at (0,-16mm) {
CONTROL PROJECT};

% Coordinator / model layer
\node[box,text width=40mm] (pc)
  at (-46mm,-39mm) {
PREPARATION\\
COORDINATOR};

\node[box,text width=32mm] (gemini)
  at (0,-39mm) {
VERTEX AI /\\
GEMINI};

\node[box,text width=40mm] (qc)
  at (46mm,-39mm) {
QUERY\\
COORDINATOR};

% Agent layer
\node[box,text width=49mm] (pa)
  at (-47mm,-66mm) {
PREPARATION AGENTS\\[0.7mm]
Schema $\cdot$ Quality $\cdot$ Transformation\\
SQL $\cdot$ Validation};

\node[box,text width=49mm] (qa)
  at (47mm,-66mm) {
ANALYTICS AGENTS\\[0.7mm]
Planner $\cdot$ SQL $\cdot$ Answer\\
Reflection};

% MCP / state layer
\node[box,text width=45mm] (pmcp)
  at (-46mm,-94mm) {
PREPARATION MCP\\[0.7mm]
Sandbox Execution $\cdot$ Validation};

\node[box,text width=32mm] (fs)
  at (0,-94mm) {
FIRESTORE\\[0.7mm]
Workflow State $\cdot$ Evidence};

\node[box,text width=45mm] (qmcp)
  at (46mm,-94mm) {
QUERY MCP\\[0.7mm]
Policy Enforcement $\cdot$ Snapshot Execution};

% ================================================================
% CONTROL-PROJECT FLOWS
% ================================================================
\draw[flow]
  (preq.south) --
  (pc.north);

\draw[flow]
  (qreq.south) --
  (qc.north);

\draw[flow]
  (pc.south) --
  node[edgeLabel]{A2A}
  (pa.north);

\draw[flow]
  (qc.south) --
  node[edgeLabel]{A2A}
  (qa.north);

\draw[flow]
  (pa.east) --
  (gemini.west);

\draw[flow]
  (qa.west) --
  (gemini.east);

\draw[flow]
  (pa.south) --
  node[edgeLabel]{MCP}
  (pmcp.north);

\draw[flow]
  (qa.south) --
  node[edgeLabel]{MCP}
  (qmcp.north);

% ================================================================
% CONTROL PROJECT BOUNDARY
% ================================================================
\begin{scope}[on background layer]
\node[
  project,
  fit=(controlTitle)(pc)(qc)(gemini)(pa)(qa)(pmcp)(qmcp)(fs),
  fill=blue!2
] (controlProj) {};
\end{scope}

% ================================================================
% CONTROL -> SANDBOX EXECUTION BOUNDARY
% ================================================================
\node[boundary,text width=45mm]
  at (-54mm,-116mm) {
CONTROL $\rightarrow$ SANDBOX\\
EXECUTION BOUNDARY};

% ================================================================
% SANDBOX PROJECT
% ================================================================
\node[title,text width=50mm] (sandboxTitle)
  at (-58mm,-132mm) {
SANDBOX PROJECT};

\node[box,text width=54mm] (sandboxBox)
  at (-58mm,-163mm) {
PER-RUN BIGQUERY SANDBOX\\[1mm]
Source Data\\
$\downarrow$\\
Generated SQL\\
$\downarrow$\\
Candidate Data Product};

\begin{scope}[on background layer]
\node[
  project,
  fit=(sandboxTitle)(sandboxBox),
  fill=orange!5
] (sandboxProj) {};
\end{scope}

% Preparation MCP -> Sandbox
\draw[flow]
  (pmcp.south)
  -- ++(0,-9mm)
  -| node[edgeLabel,pos=0.82]{execute}
  (sandboxTitle.north);

% ================================================================
% SANDBOX -> PRODUCTION AUTHORIZATION BOUNDARY
% ================================================================
\node[boundary,text width=44mm]
  at (0,-132mm) {
SANDBOX $\rightarrow$ PRODUCTION\\
AUTHORIZATION BOUNDARY};

\node[gate,text width=34mm] (promotion)
  at (0,-163mm) {
EVIDENCE-GATED\\
PRODUCTION\\
PROMOTION};

\draw[flow]
  (sandboxBox.east) --
  (promotion.west);

% ================================================================
% PRODUCTION PROJECT
% ================================================================
\node[title,text width=56mm] (prodTitle)
  at (65mm,-132mm) {
PRODUCTION PROJECT\\
BIGQUERY};

\node[box,text width=34mm] (writer)
  at (48mm,-163mm) {
PRODUCTION WRITER\\
IDENTITY\\[0.7mm]
WRITE};

\node[box,text width=34mm] (reader)
  at (84mm,-163mm) {
ANALYTICS READER\\
IDENTITY\\[0.7mm]
READ ONLY};

\node[box,text width=44mm] (prod)
  at (66mm,-190mm) {
GOVERNED\\
DATA PRODUCT};

\node[identity,text width=53mm] (idsep)
  at (66mm,-213mm) {
WRITE IDENTITY $\neq$ READ IDENTITY};

% ================================================================
% PRODUCTION PROJECT BOUNDARY
% ================================================================
\begin{scope}[on background layer]
\node[
  project,
  fit=(prodTitle)(writer)(reader)(prod)(idsep),
  fill=green!5
] (prodProj) {};
\end{scope}

% ================================================================
% PROMOTION AND PRODUCTION FLOWS
% ================================================================
\draw[flow]
  (promotion.east) --
  node[edgeLabel,above]{fenced load}
  (writer.west);

\draw[flow]
  (writer.south) --
  (prod.north west);

\draw[flow]
  (prod.north east) --
  (reader.south);

% ================================================================
% DEDICATED ANALYTICS READ CORRIDOR
% ================================================================
\coordinate (qreadTop) at (112mm,-115mm);
\coordinate (qreadEntry) at (112mm,-163mm);

\draw[flow]
  (qmcp.south)
  -- ++(0,-9mm)
  -| (qreadTop)
  --
  node[edgeLabel,rotate=90]{read-only snapshot}
  (qreadEntry)
  -- (reader.east);

% ================================================================
% SHARED ZERO-TRUST AGENT HARNESS
% ================================================================
\node[zt,text width=134mm,anchor=north] (ztbox)
  at (0,-233mm) {
ZERO-TRUST AGENT HARNESS\\[1mm]
Workload identities $\cdot$ least-privilege IAM $\cdot$ private services\\
OIDC-authenticated A2A/MCP $\cdot$ bounded execution $\cdot$ durable state};

\end{tikzpicture}
\end{adjustbox}

\caption{\textbf{Zero-trust execution architecture.}
The framework separates orchestration, sandbox execution, production promotion, and analytics access across distinct control, sandbox, and production projects. Preparation operations execute only in a Per-Run BigQuery Sandbox, and validated outputs cross the Sandbox-to-Production Authorization Boundary through a dedicated Production Writer Identity. Analytics accesses the Governed Data Product through a separate read-only Analytics Reader Identity. This separation enforces least-privilege execution, isolates generated transformations from production data, and prevents production-write authority from being shared with analytics workloads.}
\label{fig:zero-trust-agent-architecture}
\end{figure*}
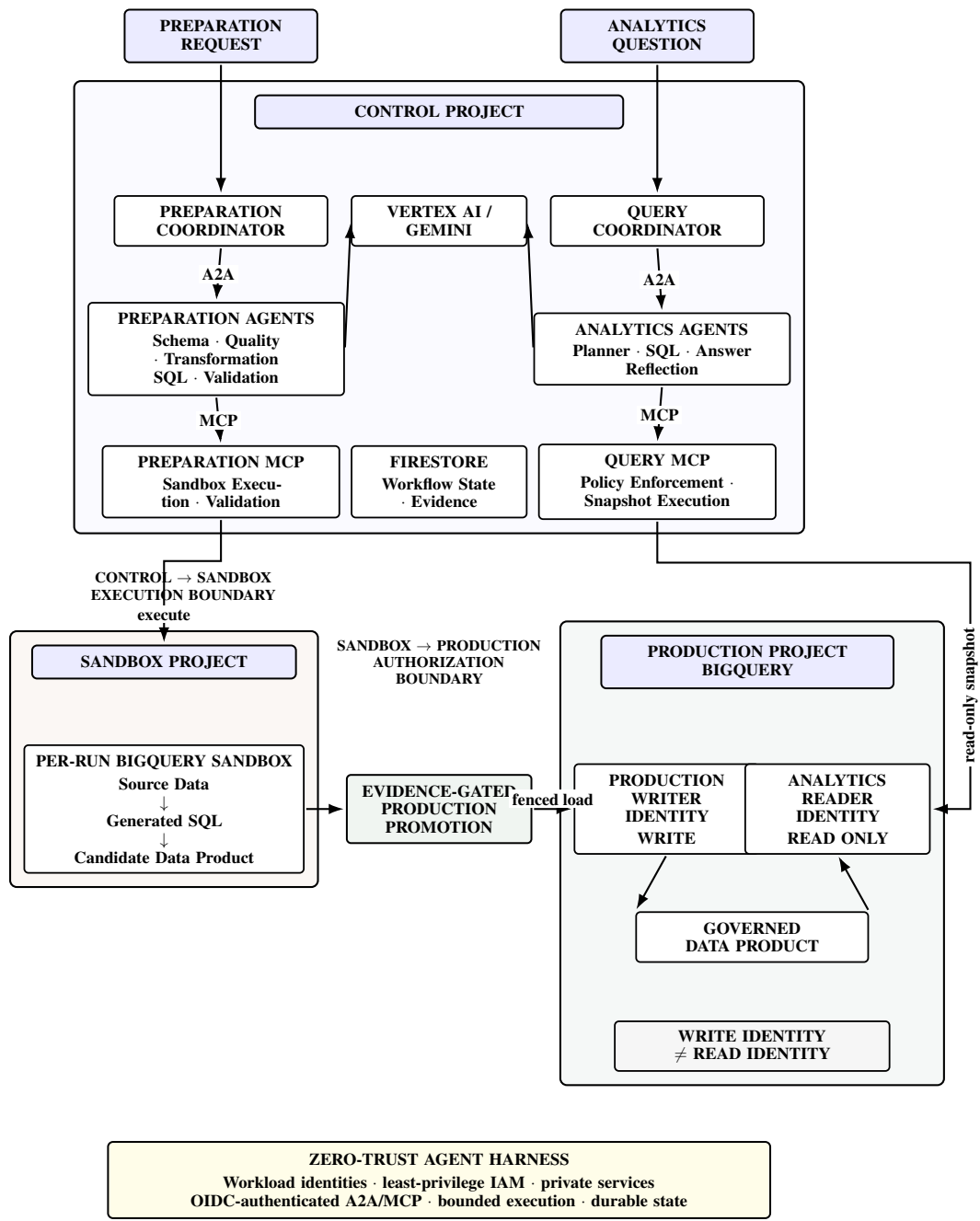

\begin{figure*}[ht!]
\centering
\begin{adjustbox}{max width=\textwidth,keepaspectratio}
\begin{tikzpicture}[
  >=Latex,
  flow/.style={-{Latex[length=2.8mm,width=1.9mm]},line width=1.05pt,
               shorten >=0.8pt,shorten <=0.8pt},
  feedback/.style={-{Latex[length=2.5mm,width=1.7mm]},dashed,line width=0.95pt,
                   shorten >=0.8pt,shorten <=0.8pt},
  edgeLabel/.style={fill=white,inner sep=1.7pt,font=\small\bfseries},
  lanehead/.style={draw,rounded corners=3pt,line width=1.1pt,fill=blue!8,
                  align=center,inner xsep=7pt,inner ysep=6pt,
                  font=\normalsize\bfseries},
  stagebox/.style={draw,rounded corners=2.5pt,line width=1.0pt,fill=white,
                  align=center,inner xsep=6pt,inner ysep=5.5pt,
                  font=\normalsize\bfseries},
  verifybox/.style={draw,rounded corners=2.5pt,line width=1.05pt,fill=green!8,
                   align=center,inner xsep=6pt,inner ysep=5.5pt,
                   font=\normalsize\bfseries},
  resultbox/.style={draw,rounded corners=2.5pt,line width=1.05pt,fill=blue!5,
                   align=center,inner xsep=6pt,inner ysep=5.5pt,
                   font=\normalsize\bfseries},
  repairbox/.style={draw,rounded corners=2.5pt,line width=1.0pt,fill=red!6,
                   align=center,inner xsep=5pt,inner ysep=5pt,
                   font=\normalsize\bfseries},
  smallbox/.style={draw,rounded corners=2pt,line width=0.95pt,fill=white,
                  align=center,inner xsep=5pt,inner ysep=5pt,
                  font=\small\bfseries},
  lane/.style={draw,rounded corners=4pt,line width=1.05pt,inner sep=8pt}
]

% ================================================================
% WORKFLOW HEADINGS
% ================================================================
\node[lanehead,text width=46mm] (prepHead)
  at (-36mm,0) {
DATA PREPARATION};

\node[lanehead,text width=58mm] (anaHead)
  at (36mm,0) {
DATA ANALYTICS};

% ================================================================
% DATA PREPARATION
% ================================================================
\node[stagebox,text width=40mm,below=10mm of prepHead] (pcoord) {
PREPARATION\\
COORDINATOR};

\node[stagebox,text width=40mm,below=10mm of pcoord] (pagents) {
PREPARATION AGENT STAGES\\[0.8mm]
Schema\\
$\downarrow$\\
Quality $\cdot$ Transformation\\
$\downarrow$\\
SQL};

\node[stagebox,text width=40mm,below=10mm of pagents] (psandbox) {
PER-RUN BIGQUERY\\
SANDBOX\\[0.8mm]
Candidate Data Product};

\node[verifybox,text width=40mm,below=10mm of psandbox] (pverify) {
PREPARATION VALIDATION\\[0.8mm]
Validation Agent\\
+ Deterministic Checks};

\node[verifybox,text width=40mm,below=10mm of pverify] (ppromotion) {
EVIDENCE-GATED\\
PRODUCTION PROMOTION};

\node[resultbox,text width=40mm,below=10mm of ppromotion] (pdata) {
GOVERNED\\
DATA PRODUCT};

% Preparation forward flow
\draw[flow]
  (prepHead.south) --
  (pcoord.north);

\draw[flow]
  (pcoord.south) --
  (pagents.north);

\draw[flow]
  (pagents.south) --
  node[edgeLabel]{execute}
  (psandbox.north);

\draw[flow]
  (psandbox.south) --
  node[edgeLabel]{validate}
  (pverify.north);

\draw[flow]
  (pverify.south) --
  node[edgeLabel]{pass}
  (ppromotion.north);

\draw[flow]
  (ppromotion.south) --
  (pdata.north);

% ================================================================
% PREPARATION BOUNDED REPAIR
% ================================================================
\node[repairbox,text width=19mm] (prepair)
  at ($(pverify.west)+(-14mm,0)$) {
BOUNDED\\
REPAIR};

\draw[feedback]
  (pverify.west) --
  node[edgeLabel,above]{fail}
  (prepair.east);

\coordinate (preturn)
  at ($(prepair.west)+(-6mm,0)$);

\draw[feedback]
  (prepair.west)
  -- (preturn)
  --
  node[edgeLabel,rotate=90]{affected stage}
  (preturn |- pagents.west)
  -- (pagents.west);

% ================================================================
% DATA ANALYTICS
% ================================================================
\node[stagebox,text width=52mm,below=10mm of anaHead] (qcoord) {
QUERY COORDINATOR};

\node[stagebox,text width=52mm,below=10mm of qcoord] (qplanner) {
QUERY PLANNER AGENT};

\node[verifybox,text width=52mm,below=10mm of qplanner] (intent) {
STRUCTURED QUERY INTENT};

% ================================================================
% INDEPENDENT QUERY CONSTRUCTION
% ================================================================
\node[smallbox,text width=27mm]
  (sqlagent)
  at ($(intent.south)+(-18mm,-17mm)$) {
ANALYTICS SQL AGENT};

\node[smallbox,text width=27mm]
  (compiler)
  at ($(intent.south)+(18mm,-17mm)$) {
CANONICAL COMPILER};

\node[smallbox,text width=27mm,below=9mm of sqlagent] (candidateQ) {
CANDIDATE QUERY};

\node[smallbox,text width=27mm,below=9mm of compiler] (canonicalQ) {
CANONICAL QUERY};

% Analytics entry flow
\draw[flow]
  (anaHead.south) --
  (qcoord.north);

\draw[flow]
  (qcoord.south) --
  (qplanner.north);

\draw[flow]
  (qplanner.south) --
  (intent.north);

% Independent paths from the same Structured Query Intent
\draw[flow]
  (intent.south)
  -- ++(0,-3mm)
  -| node[edgeLabel,pos=0.80]{model path}
  (sqlagent.north);

\draw[flow]
  (intent.south)
  -- ++(0,-3mm)
  -| node[edgeLabel,pos=0.80]{deterministic path}
  (compiler.north);

% Query construction
\draw[flow]
  (sqlagent.south) --
  (candidateQ.north);

\draw[flow]
  (compiler.south) --
  (canonicalQ.north);

% ================================================================
% SAME-SNAPSHOT EXECUTION
% ================================================================
\node[stagebox,text width=52mm]
  (qexec)
  at ($(candidateQ.south)!0.5!(canonicalQ.south)+(0,-17mm)$) {
SAME-SNAPSHOT EXECUTION};

\draw[flow]
  (candidateQ.south)
  -- ++(0,-3mm)
  -| (qexec.north west);

\draw[flow]
  (canonicalQ.south)
  -- ++(0,-3mm)
  -| (qexec.north east);

% ================================================================
% QUERY RESULTS
% ================================================================
\node[smallbox,text width=26mm]
  (candidateR)
  at ($(qexec.south)+(-16mm,-16mm)$) {
CANDIDATE\\
RESULT};

\node[smallbox,text width=26mm]
  (canonicalR)
  at ($(qexec.south)+(16mm,-16mm)$) {
CANONICAL\\
RESULT};

\draw[flow]
  (qexec.south west)
  -- ++(0,-3mm)
  -| (candidateR.north);

\draw[flow]
  (qexec.south east)
  -- ++(0,-3mm)
  -| (canonicalR.north);

% ================================================================
% RESULT VERIFICATION
% ================================================================
\node[verifybox,text width=52mm]
  (qverify)
  at ($(candidateR.south)!0.5!(canonicalR.south)+(0,-17mm)$) {
EXACT RESULT\\
EQUIVALENCE};

\draw[flow]
  (candidateR.south)
  -- ++(0,-3mm)
  -| (qverify.north west);

\draw[flow]
  (canonicalR.south)
  -- ++(0,-3mm)
  -| (qverify.north east);

% ================================================================
% ANSWER VERIFICATION
% ================================================================
\node[stagebox,text width=52mm,below=10mm of qverify] (answer) {
ANSWER AGENT};

\node[verifybox,text width=52mm,below=10mm of answer] (grounding) {
DETERMINISTIC\\
GROUNDING CHECK};

\node[stagebox,text width=52mm,below=10mm of grounding] (reflection) {
REFLECTION AGENT};

\node[resultbox,text width=52mm,below=10mm of reflection] (qresult) {
VERIFIED ANSWER};

\draw[flow]
  (qverify.south) --
  node[edgeLabel]{pass}
  (answer.north);

\draw[flow]
  (answer.south) --
  (grounding.north);

\draw[flow]
  (grounding.south) --
  node[edgeLabel]{pass}
  (reflection.north);

\draw[flow]
  (reflection.south) --
  node[edgeLabel]{pass}
  (qresult.north);

% ================================================================
% ANALYTICS BOUNDED REPAIR
% ================================================================
\node[repairbox,text width=19mm] (qrepair)
  at ($(qverify.east)+(14mm,0)$) {
BOUNDED\\
REPAIR};

\draw[feedback]
  (qverify.east) --
  node[edgeLabel,above]{fail}
  (qrepair.west);

\coordinate (qreturn)
  at ($(qrepair.east)+(6mm,0)$);

\draw[feedback]
  (qrepair.east)
  -- (qreturn)
  --
  node[edgeLabel,rotate=90]{affected stage}
  (qreturn |- qplanner.east)
  -- (qplanner.east);

% ================================================================
% BACKGROUND WORKFLOW LANES
% ================================================================
\begin{scope}[on background layer]

\node[
  lane,
  fit=(prepHead)(pcoord)(pagents)(psandbox)
      (pverify)(ppromotion)(pdata),
  fill=orange!2
] {};

\node[
  lane,
  fit=(anaHead)(qcoord)(qplanner)(intent)
      (sqlagent)(compiler)(candidateQ)(canonicalQ)
      (qexec)(candidateR)(canonicalR)(qverify)
      (answer)(grounding)(reflection)(qresult),
  fill=green!2
] {};

\end{scope}

\end{tikzpicture}
\end{adjustbox}
\caption{\textbf{Evidence-gated multi-agent workflow.} Data preparation proceeds through schema, quality, transformation, and SQL stages, followed by sandbox execution and validation before production promotion. Data analytics first converts the user question into a Structured Query Intent, from which the Candidate Query and Canonical Query are derived independently and executed against the same fixed snapshot. Exact Result Equivalence gates answer generation, after which deterministic grounding and reflection must succeed before the Verified Answer is released. Repairable failures are routed to the affected stage through bounded stage-local recovery, with at most three semantic attempts.}
\label{fig:evidence-gated-multi-agent-workflow}
\end{figure*}
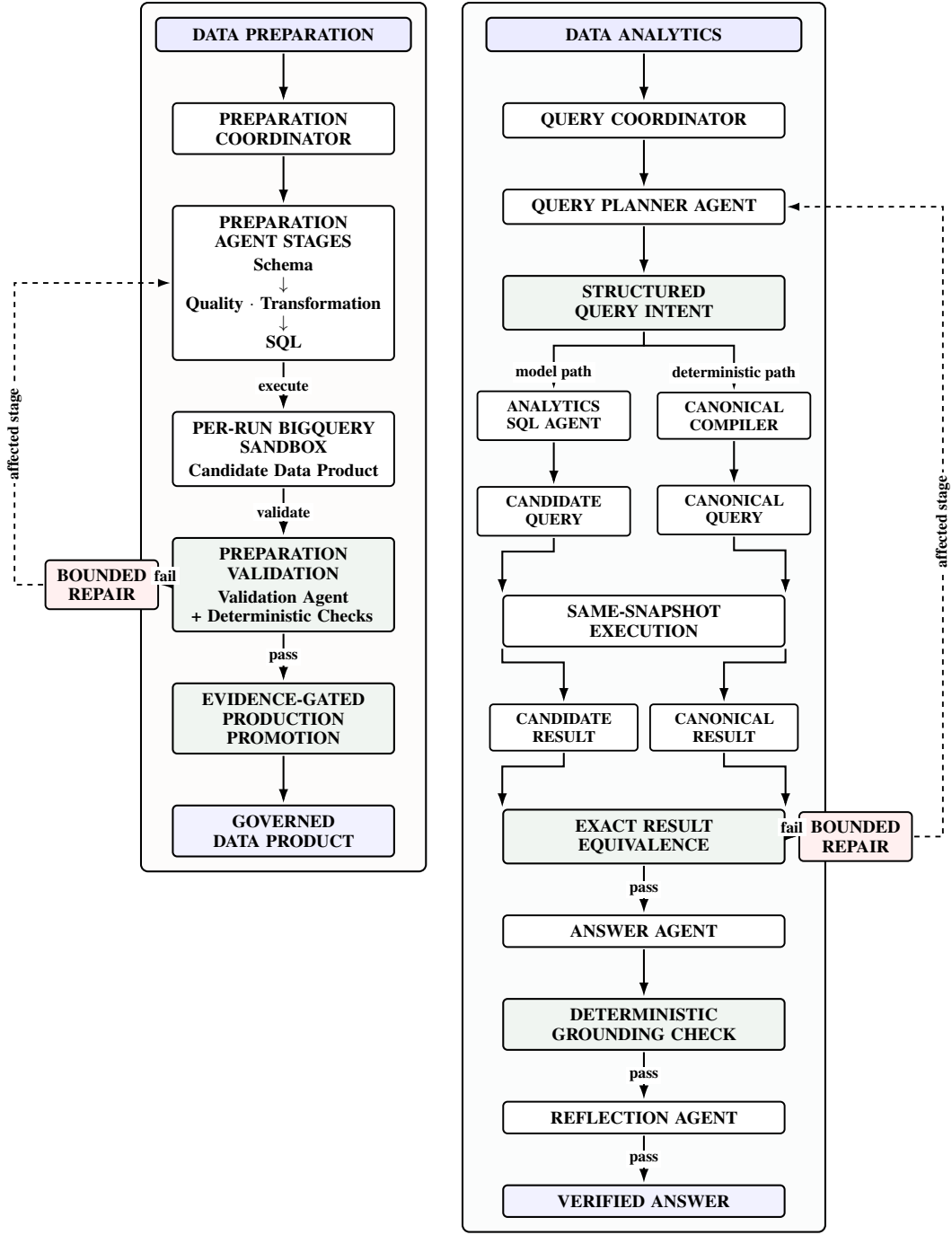

We propose \textbf{Zero-Trust Agentic OLAP}, an evidence-gated multi-agent framework that combines governed Data Preparation with verified Online Analytical Processing (OLAP). Figure~\ref{fig:overall-agentic-framework} presents the end-to-end framework, Figure~\ref{fig:zero-trust-agent-architecture} shows the execution and identity boundaries across the Control, Sandbox, and Production Projects, and
Figure~\ref{fig:evidence-gated-multi-agent-workflow} details the verification and recovery logic. The framework comprises two operationally distinct workflows: \textbf{Data Preparation}, which transforms an immutable source dataset into a validated Governed Data Product, and \textbf{OLAP}, which
answers natural-language analytical questions over that product through verified analytical execution.
(a) \textbf{Data Preparation.} A Preparation Request specifies the immutable source dataset and target data product. The Preparation Coordinator uses Google Agent Development Kit (ADK) orchestration and A2A delegation to coordinate specialist agents for schema, quality, transformation, SQL generation, and validation. Preparation MCP binds the complete source dataset to the run and executes generated SQL only within a Per-Run BigQuery Sandbox, producing a Candidate Data Product under explicit
query-byte, execution-time, and output-size bounds. The candidate is then subjected to Preparation Validation, which combines a Validation Agent with Deterministic Validation over schema conformity, key
constraints, source-to-output reconciliation, rejected-row and row-count limits, resource bounds, and execution status. Repairable failures trigger Stage-Local Bounded Repair of only the affected preparation component. Successful validation yields an Approval Binding over the immutable source,
approved preparation artifacts, executed SQL, sandbox output, and validation result, with content checksums binding the validated evidence to the promoted artifacts. Evidence-Gated Production Promotion then permits a separate least-privilege Production Writer Identity to commit the validated output as
the Governed Data Product together with its lineage and approval evidence. (b) \textbf{OLAP.}
Given a natural-language analytical question and the Governed Data Product, the Query Coordinator fixes the BigQuery snapshot and delegates the question to the Query Planner Agent. The planner produces a Structured Query Intent specifying measures, dimensions, filters, aggregations, ordering, and result semantics. From the same intent, the Analytics SQL Agent generates a Candidate Query,
while a deterministic Canonical Compiler independently derives a Canonical Query without using the Candidate Query. Query MCP enforces policy and resource constraints, mediates read-only access
to the Governed Data Product, and executes both queries against the same fixed snapshot. The Candidate Result must satisfy Exact Result Equivalence with the Canonical Result before answer generation is allowed. The Answer Agent then generates the analytical response, which must pass a Deterministic Grounding Check and Reflection Agent verification across the question, Structured Query
Intent, Candidate Query, Candidate Result, and generated answer before being returned as the Verified Answer. Repairable failures use Stage-Local Bounded Repair with a maximum of three semantic attempts.
(C) \textbf{Zero-Trust Agent Harness.} Across both workflows, sandbox transformation, production writes, and read-only OLAP access use distinct execution paths and identities. Google ADK coordinates
workflow progression, A2A carries authenticated delegation, MCP provides controlled tool and data access, and Firestore retains workflow state and execution evidence for controlled recovery. The resulting design separates model reasoning from execution authority, sandbox execution from production
authorization, and analytical generation from analytical verification.
%%%%%%%%%%%%%%%%%%%%%%%%%%%%%%%%%%%%%%%%%%%%%%%%%%%%%%%%%%%%%%%%%%%%
%%%%%%%%%%%%%%%%%%%%%%%%%%%%%%%%%%%%%%%%%%%%%%%%%%%%%%%%%%%%%%%%%%%%

\vspace{-2mm}
%%%%%%%%%%%%%%%%%%%%%%%%%%%%%%%%%%%%%%%%%%%%%%%%%%%%%%%%%%%%%%%%%%%%
%%%%%%%%%%%%%%%%%%%%%%%%%%%%%%%%%%%%%%%%%%%%%%%%%%%%%%%%%%%%%%%%%%%%
\section{Experiments}
\label{sec:experiments}
\vspace{-2mm}

\subsection{Experimental Methodology}
\label{sec:experimentalmethodology}
\vspace{-1mm}
We evaluate \textbf{Zero-Trust Agentic Data Engineering} and \textbf{Zero-Trust Agentic OLAP} independently through six research questions. For \textbf{Zero-Trust Agentic Data Engineering}, (a) \textbf{RQ1: Verified Data-Engineering System Completion} asks whether the framework can transform a natural-language data-engineering task into a complete cloud data-engineering solution and achieve verified completion only after repository, deployment, runtime, and policy verification succeeds, using bounded recovery when needed. (b) \textbf{RQ2: Data-Engineering Recovery and Termination} asks whether controlled repository, deployment, and runtime failures can be resolved through bounded diagnosis, repair, re-planning, and re-verification, and whether execution terminates correctly when the recovery budget is exhausted. (c) \textbf{RQ3: Data-Engineering Zero-Trust Execution} asks whether the agent harness denies unauthorized or out-of-scope tool and cloud operations while permitting the authorized capabilities required for generation, isolated execution, deployment, and verification. For \textbf{Zero-Trust Agentic OLAP}, (d) \textbf{RQ4: Governed Data Preparation} asks whether the framework can transform a Preparation Request and immutable source dataset into a Governed Data Product through validated, evidence-gated production promotion. (e) \textbf{RQ5: Verified OLAP} asks whether a natural-language analytical question over a Governed Data Product can produce a Verified Answer through Same-Snapshot Execution, Exact Result Equivalence, and answer grounding and verification. (f) \textbf{RQ6: OLAP Zero-Trust Execution and Bounded Recovery} asks whether verification and authorization failures prevent invalid progression while repairable failures follow Stage-Local Bounded Repair within the configured recovery bounds. Table~\ref{tab:experimental-propositions} summarizes the core experimental propositions underlying these six research questions.

\vspace{-3mm}
%%%%%%%%%%%%%%%%%%%%%%%%%%%%%%%%%%%%%%%%%%%%%%%%%%%%%%%%%%%%%%%%%%%%
\begin{table*}[ht!]
\centering
\caption{Core experimental propositions of the two proposed frameworks.}
\label{tab:experimental-propositions}
\small
\setlength{\tabcolsep}{5pt}
\renewcommand{\arraystretch}{1.18}
\begin{tabularx}{\textwidth}{
    >{\raggedright\arraybackslash}p{3.2cm}
    >{\raggedright\arraybackslash}X
}
\toprule
\textbf{Scope} & \textbf{Experimental Proposition} \\
\midrule

\textbf{Zero-Trust Agentic Data Engineering}
&
Can an LLM-driven system generate and deploy a complete cloud data-engineering solution while progression remains conditioned on repository, deployment, runtime, and policy evidence?
\\

\addlinespace

\textbf{Zero-Trust Agentic OLAP}
&
Can Data Preparation and OLAP outputs be released only after evidence-gated promotion, Same-Snapshot Execution, Exact Result Equivalence, and answer verification?
\\

\addlinespace

\textbf{Across Both Frameworks}
&
Can verification and authorization govern progression across LLMs while model capability determines successful completion and recovery within these constraints?
\\

\bottomrule
\end{tabularx}
\vspace{-3mm}
\end{table*}
%%%%%%%%%%%%%%%%%%%%%%%%%%%%%%%%%%%%%%%%%%%%%%%%%%%%%%%%%%%%%%%%%%%%

%%%%%%%%%%%%%%%%%%%%%%%%%%%%%%%%%%%%%%%%%%%%%%%%%%%%%%%%%%%%%%%%%%%%
\subsection{Benchmark Suites}
\label{sec:datasets}
\vspace{-3mm}
We construct two benchmark suites corresponding to the research questions in Section~\ref{sec:experimentalmethodology}: a \textbf{Data-Engineering Benchmark} for RQ1--RQ3 and an \textbf{OLAP Benchmark} for RQ4--RQ6. (a) \textbf{Data-Engineering Benchmark.} Tasks span representative workloads from Section~\ref{sec:introduction}, including batch ETL and ELT, real-time streaming, Change Data Capture (CDC), data lakes and lakehouses, warehouses and data marts, Internet of Things (IoT) and telemetry processing, and real-time analytics and machine-learning (ML) pipelines. Tasks vary in data sources, repository structure, pipeline and orchestration requirements, infrastructure configuration, deployment target, and runtime behavior. RQ1 uses nominal end-to-end execution. For RQ2, each task receives one repository-verification, deployment-verification, runtime-verification, or recovery-exhaustion case, distributed across the benchmark. RQ3 exercises authorized capabilities together with unauthorized or out-of-scope tool and cloud operations. (b) \textbf{OLAP Benchmark.} Each task specifies an immutable source dataset, a Preparation Request and target data product, and natural-language analytical questions over the resulting Governed Data Product. Tasks vary in schema, data-quality and transformation requirements, analytical measures, dimensions, filters, aggregations, ordering, and result semantics. RQ4 evaluates governed Data Preparation and production promotion, while RQ5 evaluates verified analytical execution and answer generation. RQ6 introduces preparation-validation failures, unauthorized production writes or OLAP reads, query-result mismatches, and answer-verification failures, with repairable cases following Stage-Local Bounded Repair. Each benchmark suite contains \textbf{100 tasks}, with every task evaluated once under each of its three RQ-aligned conditions. Across the two benchmark suites and four evaluated LLMs, this yields \textbf{2,400 task-condition executions} in total ($100$ tasks $\times$ $3$ conditions $\times$ $2$ suites $\times$ $4$ LLMs).
%%%%%%%%%%%%%%%%%%%%%%%%%%%%%%%%%%%%%%%%%%%%%%%%%%%%%%%%%%%%%%%%%%%%

%%%%%%%%%%%%%%%%%%%%%%%%%%%%%%%%%%%%%%%%%%%%%%%%%%%%%%%%%%%%%%%%%%%%
\subsection{Experimental Setup}
\label{sec:experimentalsetup}
\vspace{-1mm}
Both frameworks are implemented on Google Cloud using the execution environments described in Section~\ref{sec:overall-framework}. We evaluate four LLMs: \textbf{Gemini 2.5 Flash-Lite}, \textbf{Gemini 2.5 Flash}, \textbf{Gemini 2.5 Pro}, and \textbf{GPT-5.6 Sol}. For \textbf{Zero-Trust Agentic Data Engineering}, agents run on the regular agent-hosting Google Kubernetes Engine (GKE) layer using Google Agent Development Kit (ADK), Agent2Agent (A2A), and tenant-scoped Model Context Protocol (MCP). Repository operations use the Visual Studio Code (VS Code) Sandbox, controlled browser operations use the Chrome Sandbox, and generated workloads execute separately in a Per-Run GKE Sandbox with gVisor. Repository, deployment, runtime, policy, and audit evidence is retained for evaluation and bounded recovery. For \textbf{Zero-Trust Agentic OLAP}, Data Preparation executes in a resource-bounded Per-Run BigQuery Sandbox, with approved production promotion performed through a separate least-privilege Production Writer Identity. OLAP uses read-only Query MCP to execute Candidate and Canonical Queries against the same fixed BigQuery snapshot, while Firestore retains workflow state and execution evidence. The experiments leverage multiple \$300 Google Cloud Free Trial credits toward eligible Google Cloud charges; the only direct experimental cost beyond these credits is \textbf{GPT-5.6 Sol inference}.  For each framework, the task specification and corresponding controlled condition are held fixed across LLMs, with no model-specific prompt tuning, verification procedures, or acceptance criteria. For \textbf{Zero-Trust Agentic Data Engineering}, each retryable transition permits at most 20 correction or recovery attempts, and each execution is limited to 120 minutes of wall-clock time, 10,000,000 aggregate model tokens, and 500 tool calls. For \textbf{Zero-Trust Agentic OLAP}, Stage-Local Bounded Repair permits at most three semantic attempts for repairable analytical failures, while each execution is similarly limited to 120 minutes of wall-clock time, 10,000,000 aggregate model tokens, and 500 tool calls.
%%%%%%%%%%%%%%%%%%%%%%%%%%%%%%%%%%%%%%%%%%%%%%%%%%%%%%%%%%%%%%%%%%%%

%%%%%%%%%%%%%%%%%%%%%%%%%%%%%%%%%%%%%%%%%%%%%%%%%%%%%%%%%%%%%%%%%%%%
\subsection{Evaluation Metrics}
\label{sec:evaluationmetrics}
\vspace{-2mm}
We report one primary rate for each research question. RQ1, RQ4, and RQ5 measure successful completion of their respective nominal workflows, whereas RQ2, RQ3, and RQ6 evaluate recovery, authorization, and failure handling under controlled conditions. For RQ2, RQ3, and RQ6, each execution is judged using the success predicate corresponding to its assigned case. All metrics lie in $[0,1]$, with higher values indicating better performance. Table~\ref{tab:evaluation-metrics} summarizes the metrics.

\begin{table*}[ht!]
\vspace{-4mm}
\centering
\caption{Primary evaluation metrics for Zero-Trust Agentic Data Engineering (RQ1--RQ3) and Zero-Trust Agentic OLAP (RQ4--RQ6).}
\label{tab:evaluation-metrics}
\small
\setlength{\tabcolsep}{3.0pt}
\renewcommand{\arraystretch}{1.12}
\begin{tabularx}{\textwidth}{
    >{\raggedright\arraybackslash}p{0.50cm}
    >{\raggedright\arraybackslash}p{2.55cm}
    >{\raggedright\arraybackslash}p{3.25cm}
    X
    >{\centering\arraybackslash}p{1.10cm}
}
\toprule
\textbf{RQ}
&
\textbf{Evaluated Property}
&
\textbf{Metric}
&
\textbf{Success Criterion}
&
\textbf{Min--Max}
\\
\midrule

\multicolumn{5}{l}{\textbf{Zero-Trust Agentic Data Engineering}}\\
\midrule

RQ1
&
Verified System Completion
&
\textbf{VDECR}: Verified Data-Engineering Completion Rate
&
The task reaches verified completion with repository, deployment, runtime, and policy conditions satisfied.
&
$0$--$1$
\\

RQ2
&
Recovery and Termination
&
\textbf{DERTR}: Data-Engineering Recovery and Termination Rate
&
The assigned repository, deployment, or runtime failure is recovered and passes re-verification within budget, or a designated recovery-exhaustion case terminates correctly.
&
$0$--$1$
\\

RQ3
&
Zero-Trust Execution
&
\textbf{DEZTER}: Data-Engineering Zero-Trust Execution Rate
&
Unauthorized or out-of-scope tool and cloud operations are denied, and required authorized capabilities are permitted.
&
$0$--$1$
\\

\midrule
\multicolumn{5}{l}{\textbf{Zero-Trust Agentic OLAP}}\\
\midrule

RQ4
&
Governed Data Preparation
&
\textbf{GDPCR}: Governed Data Product Completion Rate
&
Data Preparation produces the Governed Data Product after sandboxed execution, validation, and evidence-gated production promotion.
&
$0$--$1$
\\

RQ5
&
Verified OLAP
&
\textbf{VAR}: Verified Answer Rate
&
OLAP produces a Verified Answer after Same-Snapshot Execution, Exact Result Equivalence, deterministic grounding, and reflection verification.
&
$0$--$1$
\\

RQ6
&
Zero-Trust Execution and Recovery
&
\textbf{OZTRR}: OLAP Zero-Trust and Recovery Rate
&
The applicable verification or authorization failure blocks invalid progression, and repairable cases recover and pass re-verification through Stage-Local Bounded Repair.
&
$0$--$1$
\\

\bottomrule
\end{tabularx}
\vspace{-6mm}
\end{table*}
%%%%%%%%%%%%%%%%%%%%%%%%%%%%%%%%%%%%%%%%%%%%%%%%%%%%%%%%%%%%%%%%%%%%

%%%%%%%%%%%%%%%%%%%%%%%%%%%%%%%%%%%%%%%%%%%%%%%%%%%%%%%%%%%%%%%%%%%%
\section{Results and Discussion}
\label{sec:results}
\vspace{-2mm}

\subsection{Zero-Trust Agentic Data Engineering}
\label{sec:de-results}
\vspace{-2mm}
The Data-Engineering results address RQ1--RQ3 and evaluate the framework's central hypotheses on evidence-gated completion, bounded recovery, and zero-trust execution. (a) \textbf{RQ1: Verified Data-Engineering System Completion.} As shown in Table~\ref{tab:de-main-results}, VDECR increases from 0.43 for Gemini 2.5 Flash-Lite to 0.95 for GPT-5.6 Sol, revealing a substantial cross-model gap in verified end-to-end completion. RQ1 requires repository validity, deployment verification, runtime verification,

\begin{table*}[ht!]
\centering
\caption{Primary results for Zero-Trust Agentic Data Engineering.}
\label{tab:de-main-results}
\small
\setlength{\tabcolsep}{6.0pt}
\renewcommand{\arraystretch}{1.12}
\begin{tabularx}{\textwidth}{
    >{\raggedright\arraybackslash}X
    >{\centering\arraybackslash}p{2.3cm}
    >{\centering\arraybackslash}p{2.3cm}
    >{\centering\arraybackslash}p{2.3cm}
}
\toprule
\textbf{Model}
& \textbf{VDECR}
& \textbf{DERTR}
& \textbf{DEZTER}
\\
\midrule
Gemini 2.5 Flash-Lite & 0.43 & 0.56 & 0.98 \\
Gemini 2.5 Flash      & 0.55 & 0.66 & 0.99 \\
Gemini 2.5 Pro        & 0.69 & 0.76 & 0.99 \\
GPT-5.6 Sol           & 0.95 & 0.97 & 1.00 \\
\bottomrule
\end{tabularx}
\vspace{-5mm}
\end{table*}

and policy compliance to be jointly satisfied before terminal success is permitted. The results support the hypothesis underlying evidence-gated completion: complete cloud data-engineering solutions can reach verified completion under independently established verification conditions, while satisfying the full end-to-end workflow remains strongly dependent on model capability. \textbf{The key novelty is that terminal success is conditioned on retained repository, deployment, runtime, and policy evidence rather than on generation or agent-reported success alone.} (b) \textbf{RQ3: Data-Engineering Zero-Trust Execution.} Table~\ref{tab:de-main-results} shows that DEZTER remains between 0.98 and 1.00 across all models, substantially more stable than VDECR and DERTR. RQ3 requires unauthorized or out-of-scope tool and cloud operations to be denied while permitting the authorized capabilities required for generation, isolated execution, deployment, and verification. The near-ceiling results support the hypothesis that zero-trust execution can be enforced largely independently of model capability. \textbf{The principal finding is the separation of model reasoning from execution authority:} agents perform generation and recovery, whereas workload identity, authorization, policy enforcement, MCP-mediated tool access, and isolated execution constrain consequential operations.

\begin{table*}[ht!]
\vspace{-2mm}
\centering
\caption{Condition-level outcomes for controlled Data-Engineering failure cases.}
\label{tab:de-recovery-breakdown}
\small
\setlength{\tabcolsep}{4.0pt}
\renewcommand{\arraystretch}{1.10}
\begin{tabularx}{\textwidth}{
    >{\raggedright\arraybackslash}X
    >{\centering\arraybackslash}p{2.05cm}
    >{\centering\arraybackslash}p{2.05cm}
    >{\centering\arraybackslash}p{2.05cm}
    >{\centering\arraybackslash}p{2.05cm}
}
\toprule
\textbf{Controlled Condition}
& \textbf{Flash-Lite}
& \textbf{Flash}
& \textbf{Pro}
& \textbf{GPT-5.6 Sol}
\\
\midrule
Repository-verification repair
& 0.32 & 0.48 & 0.60 & 0.92 \\
Deployment-verification repair
& 0.52 & 0.60 & 0.76 & 1.00 \\
Runtime-verification repair
& 0.40 & 0.56 & 0.68 & 0.96 \\
Recovery-exhaustion termination
& 1.00 & 1.00 & 1.00 & 1.00 \\
\bottomrule
\end{tabularx}
\vspace{-2mm}
\end{table*}

(c) \textbf{RQ2: Data-Engineering Recovery and Termination.}
Table~\ref{tab:de-main-results} shows that DERTR ranges from 0.56 for Flash-Lite to 0.97 for GPT-5.6 Sol, while Table~\ref{tab:de-recovery-breakdown} identifies the corresponding condition-level behavior. Repository-verification repair is the most difficult repair condition across all models, whereas deployment-verification repair achieves the highest rate. Repository failures may require coordinated changes across code, configuration, dependencies, and infrastructure definitions, while deployment failures can expose comparatively structured deployment diagnostics. Runtime recovery additionally requires reasoning over runtime evidence, including application health, API responses, service interactions, pipeline execution, logs, metrics, and traces. Recovery-exhaustion termination reaches 1.00 for every model. These results support the hypothesis underlying verification-driven bounded recovery: repository, deployment, and runtime failures can enter bounded diagnosis, repair, re-planning, and re-verification, while exhaustion of the recovery budget produces correct termination. \textbf{The central result is that repair success remains model-dependent, whereas bounded termination is consistently enforced by the framework.} Please see the technical appendix for additional results and discussion.
%%%%%%%%%%%%%%%%%%%%%%%%%%%%%%%%%%%%%%%%%%%%%%%%%%%%%%%%%%%%%%%%%%%%

\vspace{-3mm}
%%%%%%%%%%%%%%%%%%%%%%%%%%%%%%%%%%%%%%%%%%%%%%%%%%%%%%%%%%%%%%%%%%%%
\section{Conclusion}
\label{sec:conclusion}
\vspace{-3mm}
We presented \textbf{Zero-Trust Agentic Data Engineering} and \textbf{Zero-Trust Agentic OLAP}, two complementary frameworks for verifiable agentic data workflows. Zero-Trust Agentic Data Engineering generates, deploys, and verifies complete cloud data-engineering solutions, while Zero-Trust Agentic OLAP combines governed Data Preparation with verified analytical execution over Governed Data Products. Both frameworks instantiate \textbf{graph engineering}, \textbf{loop engineering}, and \textbf{agent-harness engineering} for evidence-gated progression, bounded recovery, and zero-trust execution. Data Engineering reaches completion only after repository, deployment, runtime, and policy verification, whereas OLAP permits production promotion only after Preparation Validation and releases a Verified Answer only after Same-Snapshot Execution, Exact Result Equivalence, deterministic grounding, and reflection. Together, these mechanisms couple autonomous generation and repair with isolated execution, explicit authorization, and independent verification, so that terminal success is determined by retained evidence rather than agent-reported success.
%%%%%%%%%%%%%%%%%%%%%%%%%%%%%%%%%%%%%%%%%%%%%%%%%%%%%%%%%%%%%%%%%%%%

\clearpage
\newpage

%====================================%
% References 
%====================================% 
\nocite{*}
\bibliographystyle{plainnat}
\bibliography{references}

%====================================%
% Technical Appendix 
%====================================%
\clearpage

\section{Technical Appendix}
\label{sec:technical_appendix}

%%%%%%%%%%%%%%%%%%%%%%%%%%%%%%%%%%%%%%%%%%%%%%%%%%%%%%%%%%%%%%%%%%%%
\subsection{Zero-Trust Agentic OLAP}
\label{sec:olap-results}
The Zero-Trust Agentic OLAP results address RQ4--RQ6 and evaluate governed Data Preparation, verified OLAP, and zero-trust execution with bounded recovery.

\begin{table*}[ht!]
\centering
\caption{Primary results for Zero-Trust Agentic OLAP.}
\label{tab:olap-main-results}
\small
\setlength{\tabcolsep}{6pt}
\renewcommand{\arraystretch}{1.12}

\begin{tabularx}{\textwidth}{
    >{\raggedright\arraybackslash}X
    >{\centering\arraybackslash}p{2.3cm}
    >{\centering\arraybackslash}p{2.3cm}
    >{\centering\arraybackslash}p{2.3cm}
}
\toprule
\textbf{Model} &
\textbf{GDPCR} &
\textbf{VAR} &
\textbf{OZTRR} \\
\midrule
Gemini 2.5 Flash-Lite & 0.62 & 0.47 & 0.66 \\
Gemini 2.5 Flash      & 0.71 & 0.58 & 0.74 \\
Gemini 2.5 Pro        & 0.81 & 0.70 & 0.82 \\
GPT-5.6 Sol           & 0.96 & 0.94 & 0.97 \\
\bottomrule
\end{tabularx}
\end{table*}

(a) \textbf{RQ4: Governed Data Preparation.} As shown in Table~\ref{tab:olap-main-results}, GDPCR increases from 0.62 for Gemini 2.5 Flash-Lite to 0.96 for GPT-5.6 Sol, with Flash and Pro reaching 0.71 and 0.81, respectively. Successful completion requires sandboxed preparation, deterministic and agent-based validation, evidence binding, and Evidence-Gated Production Promotion rather than successful SQL execution alone. \textbf{The key result is that production promotion is conditioned on independently established validation evidence, while successful creation of the Governed Data Product remains model-dependent.} (b) \textbf{RQ5: Verified OLAP.} VAR increases from 0.47 for Flash-Lite to 0.94 for GPT-5.6 Sol. The Candidate Query is verified against an independently derived Canonical Query through Same-Snapshot Execution and Exact Result Equivalence before answer generation. Consequently, semantically incorrect queries fail verification even when they are syntactically executable. Deterministic grounding and reflection additionally verify consistency of the generated answer with the computed result. \textbf{The key result is that OLAP correctness extends beyond executable SQL to independent verification of both the computed result and the grounded analytical answer.}

\begin{table*}[ht!]
\centering
\caption{Condition-level outcomes for controlled Zero-Trust Agentic OLAP failure cases.}
\label{tab:olap-recovery-breakdown}
\small
\setlength{\tabcolsep}{4pt}
\renewcommand{\arraystretch}{1.10}

\begin{tabularx}{\textwidth}{
    >{\raggedright\arraybackslash}X
    >{\centering\arraybackslash}p{2.05cm}
    >{\centering\arraybackslash}p{2.05cm}
    >{\centering\arraybackslash}p{2.05cm}
    >{\centering\arraybackslash}p{2.05cm}
}
\toprule
\textbf{Controlled Condition} &
\textbf{Flash-Lite} &
\textbf{Flash} &
\textbf{Pro} &
\textbf{GPT-5.6 Sol} \\
\midrule
Preparation-validation repair &
0.55 & 0.65 & 0.80 & 1.00 \\

Unauthorized production write / OLAP read denied &
1.00 & 1.00 & 1.00 & 1.00 \\

Query-result mismatch repair &
0.30 & 0.45 & 0.60 & 0.90 \\

Answer-verification repair &
0.45 & 0.60 & 0.70 & 0.95 \\
\bottomrule
\end{tabularx}
\end{table*}

(c) \textbf{RQ6: OLAP Zero-Trust Execution and Bounded Recovery.} As shown in Table~\ref{tab:olap-main-results}, OZTRR ranges from 0.66 for Flash-Lite to 0.97 for GPT-5.6 Sol. Table~\ref{tab:olap-recovery-breakdown} shows that query-result mismatch repair is the most difficult condition, followed by answer-verification repair, whereas preparation-validation repair is comparatively easier. Repairable failures remain confined to Stage-Local Bounded Repair, while unauthorized production writes and OLAP reads are denied for every model through separated execution paths and identities. \textbf{The key result is the separation of model-dependent recovery capability from framework-enforced authorization and prevention of invalid workflow progression.} Please see the technical appendix for additional results and discussion.
%%%%%%%%%%%%%%%%%%%%%%%%%%%%%%%%%%%%%%%%%%%%%%%%%%%%%%%%%%%%%%%%%%%%

% %%%%%%%%%%%%%%%%%%%%%%%%%%%%%%%%%%%%%%%%%%%%%%%%%%%%%%%%%%%%%%%%%%%%
%%%%%%%%%%%%%%%%%%%%%%%%%%%%%%%%%%%%%%%%%%%%%%%%%%%%%%%%%%%%%%%%%%%%
\subsection{Ablation Studies}
\label{sec:ablations}
\vspace{-2mm}
\subsubsection{Ablation Protocol}
We ablate three shared control mechanisms corresponding to the proposed framework abstractions: \textbf{evidence-gated progression} in graph engineering, \textbf{bounded recovery} in loop engineering, and \textbf{zero-trust authorization and policy enforcement} in agent-harness engineering. Each configuration uses the same 600 task-condition executions per LLM across the Data-Engineering and OLAP benchmarks, with the original verification, authorization, recovery, and terminal success criteria retained for scoring. Removing evidence-gated progression allows forward transitions despite unsatisfied verification predicates. Removing bounded recovery disables diagnosis, repair, re-planning, retry, and re-verification in Data Engineering and Stage-Local Bounded Repair in OLAP. Removing zero-trust authorization and policy enforcement disables online permit/deny enforcement while retaining the expected decisions for evaluation. The primary evaluation contains 2,400 executions; the three ablations add 7,200, yielding 9,600 executions in total.

\begin{table*}[ht!]
\centering
\caption{Ablation of shared control mechanisms using the primary evaluation metrics.}
\label{tab:shared-ablation}
\scriptsize
\setlength{\tabcolsep}{2.5pt}
\renewcommand{\arraystretch}{1.08}

\begin{tabularx}{\textwidth}{X l c c c c c c}
\toprule
\textbf{Configuration} & \textbf{Model} & \textbf{VDECR} & \textbf{DERTR} & \textbf{DEZTER} & \textbf{GDPCR} & \textbf{VAR} & \textbf{OZTRR} \tabularnewline
\midrule

Full framework & Flash-Lite & 0.43 & 0.56 & 0.98 & 0.62 & 0.47 & 0.66 \tabularnewline
& Flash & 0.55 & 0.66 & 0.99 & 0.71 & 0.58 & 0.74 \tabularnewline
& Pro & 0.69 & 0.76 & 0.99 & 0.81 & 0.70 & 0.82 \tabularnewline
& GPT-5.6 Sol & 0.95 & 0.97 & 1.00 & 0.96 & 0.94 & 0.97 \tabularnewline

\midrule
w/o evidence-gated progression & Flash-Lite & 0.31 & 0.45 & 0.97 & 0.48 & 0.36 & 0.54 \tabularnewline
& Flash & 0.41 & 0.54 & 0.99 & 0.56 & 0.45 & 0.62 \tabularnewline
& Pro & 0.53 & 0.65 & 0.98 & 0.68 & 0.57 & 0.71 \tabularnewline
& GPT-5.6 Sol & 0.78 & 0.85 & 1.00 & 0.82 & 0.80 & 0.86 \tabularnewline

\midrule
w/o bounded recovery & Flash-Lite & 0.28 & 0.21 & 0.97 & 0.44 & 0.31 & 0.29 \tabularnewline
& Flash & 0.38 & 0.29 & 0.98 & 0.52 & 0.40 & 0.38 \tabularnewline
& Pro & 0.50 & 0.38 & 0.99 & 0.63 & 0.50 & 0.47 \tabularnewline
& GPT-5.6 Sol & 0.58 & 0.45 & 1.00 & 0.71 & 0.62 & 0.56 \tabularnewline

\midrule
w/o zero-trust authorization / policy & Flash-Lite & 0.41 & 0.54 & 0.45 & 0.60 & 0.45 & 0.51 \tabularnewline
& Flash & 0.53 & 0.64 & 0.52 & 0.69 & 0.56 & 0.58 \tabularnewline
& Pro & 0.67 & 0.74 & 0.58 & 0.79 & 0.68 & 0.65 \tabularnewline
& GPT-5.6 Sol & 0.93 & 0.96 & 0.61 & 0.95 & 0.93 & 0.82 \tabularnewline

\bottomrule
\end{tabularx}
\end{table*}

(a) \textbf{Evidence-Gated Progression.} Removing evidence-gated progression reduces VDECR by 0.12--0.17, GDPCR by 0.13--0.15, and VAR by 0.11--0.14 across models. DERTR and OZTRR also decrease by 0.11--0.12 because execution can advance beyond unsatisfied verification predicates, whereas DEZTER remains near ceiling at 0.97--1.00 with authorization enforcement unchanged. \textbf{The results show that evidence-gated graph progression contributes to verified completion and recovery by preventing invalid forward transitions.} (b) \textbf{Bounded Recovery.} Removing bounded recovery produces the largest decreases in DERTR (0.35--0.52) and OZTRR (0.35--0.41). VDECR, GDPCR, and VAR also decline by 0.15--0.37, 0.18--0.25, and 0.16--0.32, respectively, because repairable failures can no longer be corrected and re-verified. GPT-5.6 Sol shows the largest absolute decreases in these completion metrics: VDECR from 0.95 to 0.58, GDPCR from 0.96 to 0.71, and VAR from 0.94 to 0.62. DEZTER remains at 0.97--1.00. \textbf{The results show that loop engineering is critical for converting repairable failures into re-verified completion while leaving authorization enforcement largely unchanged.} (c) \textbf{Zero-Trust Authorization and Policy Enforcement.} Removing online authorization and policy enforcement causes the dominant degradation in DEZTER, from 0.98 to 0.45 for Flash-Lite, 0.99 to 0.52 for Flash, 0.99 to 0.58 for Pro, and 1.00 to 0.61 for GPT-5.6 Sol. OZTRR decreases by 0.15--0.17 because RQ6 includes unauthorized production-write and OLAP-read conditions. Residual DEZTER reflects cases in which attempted operations remain blocked by other active execution constraints or are not issued during the run. In contrast, VDECR, DERTR, GDPCR, and VAR remain within 0.01--0.02 of their full-framework values. \textbf{The results show that authorization and policy enforcement primarily govern execution authority while having little effect on measured generation, verification, and recovery performance.} 
Overall, Table~\ref{tab:shared-ablation} shows complementary effects across the three mechanisms: \textbf{graph engineering governs verified progression, loop engineering governs bounded recovery, and agent-harness authorization and policy enforcement govern execution authority.}

% %%%%%%%%%%%%%%%%%%%%%%%%%%%%%%%%%%%%%%%%%%%%%%%%%%%%%%%%%%%%%%%%%%%%
% %%%%%%%%%%%%%%%%%%%%%%%%%%%%%%%%%%%%%%%%%%%%%%%%%%%%%%%%%%%%%%%%%%%%

%%%%%%%%%%%%%%%%%%%%%%%%%%%%%%%%%%%%%%%%%%%%%%%%%%%%%%%%%%%%%%%%%%%%
%%%%%%%%%%%%%%%%%%%%%%%%%%%%%%%%%%%%%%%%%%%%%%%%%%%%%%%%%%%%%%%%%%%%
\subsection{Representative Benchmark Tasks}
\label{sec}
\vspace{-1mm}
Tables~\ref{tab:de-representative-tasks} and~\ref{tab:olap-representative-tasks} show representative tasks from the two 100-task benchmark suites. All tasks use fixed, checksummed snapshots, subsets, or deterministic replay sequences derived from publicly available datasets and held constant across LLMs and executions. (a) For the \textbf{Data-Engineering Benchmark}, Table~\ref{tab:de-representative-tasks} illustrates tasks spanning mobility, transportation, e-commerce, recommendation, software-development, streaming-style, and multi-source analytics. Dataset, application, and level are benchmark annotations; the framework receives only the end-user natural-language cloud task and its run-bound data source. It must determine the architecture, generate the complete repository and Infrastructure as Code, validate and execute the generated artifacts in isolation, deploy the cloud solution, and satisfy repository, deployment, runtime, and policy verification before completion. RQ2 perturbations are introduced independently of the nominal task and are not encoded in the user request. Easy, Medium, and Hard levels reflect increasing system and workflow complexity, including source integration, generated components, execution mode, statefulness, temporal processing, orchestration, and runtime-serving requirements, rather than dataset size alone.
(b) For the \textbf{OLAP Benchmark}, Table~\ref{tab:olap-representative-tasks} presents tasks consisting of an immutable benchmark source, a natural-language Preparation Request, and a natural-language analytical question. The Preparation Request defines the required semantics of the Governed Data Product without prescribing implementation SQL, execution strategy, or verification procedures, while the analytical question defines the measures, dimensions, filters, temporal semantics, rankings, and result semantics required for independent Canonical Query construction and Exact Result Equivalence. The framework performs sandboxed Data Preparation, validation, Evidence-Gated Production Promotion, Same-Snapshot OLAP execution, result verification, deterministic grounding, and answer verification. Easy tasks emphasize validation and basic aggregation, Medium tasks add richer transformations, derived measures, and multidimensional analysis, and Hard tasks introduce temporal comparisons, ranking or window semantics, multiple dimensions, and cross-source analysis. RQ6 failures are applied separately from the nominal Preparation Request and analytical question.

%%%%%%%%%%%%%%%%%%%%%%%%%%%%%%%%%%%%%%%%%%%%%%%%%%%%%%%%%%%%%%%%%%%%
% REPRESENTATIVE DATA-ENGINEERING TASKS
%%%%%%%%%%%%%%%%%%%%%%%%%%%%%%%%%%%%%%%%%%%%%%%%%%%%%%%%%%%%%%%%%%%%
\begin{table*}[ht!]
\centering
\caption{Representative tasks from the 100-task Data-Engineering Benchmark. Level denotes task complexity rather than dataset complexity. Dataset, application, and level are benchmark annotations; the framework receives only the end-user natural-language cloud task.}
\label{tab:de-representative-tasks}
\small
\setlength{\tabcolsep}{3.5pt}
\renewcommand{\arraystretch}{1.12}
\begin{tabularx}{\textwidth}{
    >{\raggedright\arraybackslash}p{0.9cm}
    >{\raggedright\arraybackslash}p{2.3cm}
    >{\raggedright\arraybackslash}p{2.3cm}
    X
}
\toprule
\textbf{Level} &
\textbf{Dataset} &
\textbf{Application} &
\textbf{Representative End-User Natural-Language Cloud Task}
\tabularnewline
\midrule

Easy &
NYC Citi Bike &
Mobility analytics &
Using the benchmark-provided Citi Bike trip-data snapshot, build and deploy a cloud application that validates and processes trip records, maintains analytics-ready trip and station data, and provides APIs and dashboards for trip departures, station activity, ride duration, and usage patterns over time.
\tabularnewline

% Easy &
% Stack Overflow &
% Developer analytics &
% Using the benchmark-provided Stack Overflow data snapshot, build and deploy a cloud platform that validates and processes question and related post data and provides APIs and a dashboard for analyzing question volume, tags, views, scores, answers, and activity over time.
% \tabularnewline

Medium &
Chicago Taxi Trips &
Transportation analytics &
Using the benchmark-provided Chicago Taxi Trips snapshot delivered as deterministic incremental batches, build and deploy a transportation data platform that incrementally validates and processes trip data, maintains consistent analytical datasets, and provides APIs and dashboards for trip demand, fares, distance, payment type, and geographic and temporal trends.
\tabularnewline

Medium &
GA4 Sample E-Commerce &
E-commerce analytics &
Using the benchmark-provided GA4 Sample E-Commerce snapshot, build and deploy an e-commerce data platform that processes event, traffic-source, device, product, session, and purchase information, maintains curated analytical datasets, and provides APIs and dashboards for revenue, purchase-session conversion, products, devices, traffic sources, and customer activity.
\tabularnewline

% Medium &
% MovieLens 32M &
% Recommendation analytics &
% Using the benchmark-provided MovieLens 32M snapshot, build and deploy a cloud data platform that integrates movie, rating, and tag data, validates their relationships, maintains analytics- and recommendation-ready datasets, and provides APIs and dashboards for ratings, genres, movies, tags, user interactions, and recommendation-oriented features.
% \tabularnewline

Medium &
GH Archive &
Software-development analytics &
Using the benchmark-provided ordered sequence of GH Archive files, build and deploy a cloud platform that incrementally processes public GitHub activity events and maintains curated repository-activity data, APIs, and dashboards for commits, issues, pull requests, forks, and other development activity over time.
\tabularnewline

Hard &
NYC Citi Bike &
Near-real-time mobility platform &
Using the benchmark-provided deterministic event-stream replay of Citi Bike trip records, build and deploy a near-real-time mobility platform that handles delayed and duplicate events, performs stateful incremental processing, continuously updates validated analytical data, and provides current station and trip metrics through backend APIs and dashboards.
\tabularnewline

Hard &
GA4 Sample E-Commerce &
Resilient event-processing platform &
Using the benchmark-provided deterministic event-stream replay of GA4 e-commerce events, build and deploy a production-oriented event-processing platform that performs idempotent incremental processing, maintains validated session, product, purchase, revenue, and conversion data, detects data-quality and freshness problems, and serves current business metrics through APIs and dashboards.
\tabularnewline

Hard &
NOAA GSOD + Chicago Taxi Trips &
Multi-source transportation analytics &
Using the benchmark-provided NOAA GSOD and Chicago Taxi Trips snapshots, build and deploy a cloud data platform that validates both sources, associates taxi pickup areas with eligible weather stations through the benchmark-defined geographic mapping, aligns trips with daily weather observations by date, maintains integrated analytical datasets, and provides APIs and dashboards for analyzing relationships among weather, trip demand, travel time, distance, and fares.
\tabularnewline

Hard &
Stack Overflow + GH Archive &
Engineering analytics platform &
Using the benchmark-provided Stack Overflow and GH Archive snapshots, build and deploy an end-to-end cloud data platform that independently processes both sources, aligns their activity by benchmark-defined technology categories and calendar periods, continuously maintains validated analytical datasets, monitors pipeline health and data quality, and provides backend APIs and a frontend for multidimensional developer-community and software-development analytics.
\tabularnewline

\bottomrule
\end{tabularx}
\end{table*}

%%%%%%%%%%%%%%%%%%%%%%%%%%%%%%%%%%%%%%%%%%%%%%%%%%%%%%%%%%%%%%%%%%%%
% REPRESENTATIVE OLAP TASKS
%%%%%%%%%%%%%%%%%%%%%%%%%%%%%%%%%%%%%%%%%%%%%%%%%%%%%%%%%%%%%%%%%%%%
\begin{table*}[ht!]
\centering
\caption{Representative tasks from the 100-task OLAP Benchmark. Each task contains an immutable benchmark snapshot derived from a publicly available source dataset, a natural-language Preparation Request, and a natural-language analytical question. Level denotes preparation and analytical complexity rather than dataset complexity.}
\label{tab:olap-representative-tasks}
\small
\setlength{\tabcolsep}{3.2pt}
\renewcommand{\arraystretch}{1.12}
\begin{tabularx}{\textwidth}{
>{\raggedright\arraybackslash}p{0.85cm}
>{\raggedright\arraybackslash}p{2.15cm}
>{\raggedright\arraybackslash}p{5.0cm}
X
}
\toprule
\textbf{Level} &
\textbf{Dataset} &
\textbf{Preparation Request} &
\textbf{Representative OLAP Question}
\tabularnewline
\midrule

% Easy &
% NOAA GSOD &
% Prepare a governed weather data product containing valid station, date,
% temperature, precipitation, and wind attributes, with invalid or missing
% observations handled consistently. &
% What are the average temperature and total precipitation by station and month?
% \tabularnewline

% Easy &
% NYC Citi Bike &
% Prepare a governed trip data product with validated trip identifiers, start and end times, start and end stations, derived ride duration, and available rider attributes; remove duplicate trip identifiers and records with invalid temporal or station information. &
% For each calendar month, which start stations have the highest number of trip departures?
% \tabularnewline

Easy &
Stack Overflow &
Prepare a governed question data product with validated question identifiers, creation dates, scores, answer counts, and view counts, and normalize tags so that each distinct question--tag relationship is represented once. &
For each year, which tags have the highest number of distinct questions, and what is the mean view count per question for each of those tags?
\tabularnewline

Medium &
Chicago Taxi Trips &
Prepare a governed taxi-trip data product with validated timestamps, trip duration, distance, fare, payment type, and geographic attributes; exclude invalid nonpositive duration or distance values where required by derived measures and derive consistent month and hour-of-day dimensions. &
How do trip count, mean fare, and mean trip distance vary by payment type, calendar month, and pickup hour of day?
\tabularnewline

% Medium &
% GA4 Sample E-Commerce &
% Prepare a governed e-commerce data product from event records by validating
% events, extracting relevant product and purchase attributes, identifying
% sessions, and deriving revenue and conversion measures. &
% How do purchase revenue and conversion rate vary by traffic source, device
% category, and month?
% \tabularnewline

Medium &
MovieLens 32M &
Prepare a governed movie-rating data product by validating movies, ratings, and tags, checking movie--rating relationships, expanding each movie into its listed genres, and deriving each movie's rating count and mean rating. &
Among movies with at least 1,000 ratings, which genres have the highest mean of the qualifying movies' movie-level average ratings?
\tabularnewline

% Medium &
% GH Archive &
% Prepare a governed repository-activity data product containing validated repository identifier, actor, event type, and event timestamp information, and derive monthly event counts separately for issue, pull-request, and fork events. &
% For each calendar month, which repositories have the highest counts of issue events, pull-request events, and fork events?
% \tabularnewline

Hard &
NYC Citi Bike &
Prepare a governed mobility data product with validated station, trip, and duration attributes; derive local calendar date and time dimensions, define weekday morning as Monday--Friday from 06:00 through 11:59, and derive calendar quarter and meteorological season for each trip. &
For each calendar quarter and meteorological season, which start stations rank highest by weekday-morning trip departures, and how does each station's rank change between consecutive seasons?
\tabularnewline

Hard &
GA4 Sample E-Commerce &
Prepare a governed commerce data product with validated users, traffic sources, device categories, products, purchases, and purchase revenue; derive sessions from the benchmark-defined user and session identifiers and define purchase-session conversion rate as the number of sessions containing at least one purchase divided by the total number of sessions in the corresponding group. &
For each calendar month, which traffic-source, device-category, and product-category combinations have the highest purchase-session conversion rate and total purchase revenue, and how do both measures change from the preceding month?
\tabularnewline

Hard &
NOAA GSOD + Chicago Taxi Trips &
Prepare a governed analytical data product from the benchmark-provided weather and taxi snapshots by associating each taxi pickup area with its benchmark-defined eligible NOAA station, joining trips to the corresponding daily observation by local date, converting temperature to degrees Celsius, deriving fixed 5-degree-Celsius temperature bands and a precipitation-present indicator, and excluding records with invalid nonpositive trip distance when deriving fare per mile. &
By calendar month, pickup area, 5-degree-Celsius temperature band, and precipitation-present status, how do taxi-trip count, mean trip duration, and mean fare per mile vary?
\tabularnewline

Hard &
Stack Overflow &
Prepare a governed question-and-answer data product with validated questions, creation year, tags, answer counts, and view counts; normalize question--tag relationships, define answer rate as the fraction of questions with at least one answer, and derive year-over-year percentage change in distinct question count for each tag. &
For each year, rank technology tags separately by year-over-year question-count growth, answer rate, and mean views per question, and identify the five tags with the largest year-over-year decline in question count.
\tabularnewline

\bottomrule
\end{tabularx}
\end{table*}

%%%%%%%%%%%%%%%%%%%%%%%%%%%%%%%%%%%%%%%%%%%%%%%%%%%%%%%%%%%%%%%%%%%%
%%%%%%%%%%%%%%%%%%%%%%%%%%%%%%%%%%%%%%%%%%%%%%%%%%%%%%%%%%%%%%%%%%%%

%%%%%%%%%%%%%%%%%%%%%%%%%%%%%%%%%%%%%%%%%%%%%%%%%%%%%%%%%%%%%%%%%%%%
%%%%%%%%%%%%%%%%%%%%%%%%%%%%%%%%%%%%%%%%%%%%%%%%%%%%%%%%%%%%%%%%%%%%
% FIGURE 2 -- ZERO-TRUST EXECUTION ARCHITECTURE
% %%%%%%%%%%%%%%%%%%%%%%%%%%%%%%%%%%%%%%%%%%%%%%%%%%%%%%%%%%%%%%%%%%%%%%%%%%
% %%%%%%%%%%%%%%%%%%%%%%%%%%%%%%%%%%%%%%%%%%%%%%%%%%%%%%%%%%%%%%%%%%%%%%%%%%

\clearpage

% %%%%%%%%%%%%%%%%%%%%%%%%%%%%%%%%%%%%%%%%%%%%%%%%%%%%%%%%%%%%%%%%%%%%%%%%%%
% %%%%%%%%%%%%%%%%%%%%%%%%%%%%%%%%%%%%%%%%%%%%%%%%%%%%%%%%%%%%%%%%%%%%%%%%%%
% FIGURE 3 -- EVIDENCE-GATED MULTI-AGENT WORKFLOW
% %%%%%%%%%%%%%%%%%%%%%%%%%%%%%%%%%%%%%%%%%%%%%%%%%%%%%%%%%%%%%%%%%%%%%%%%%%
% %%%%%%%%%%%%%%%%%%%%%%%%%%%%%%%%%%%%%%%%%%%%%%%%%%%%%%%%%%%%%%%%%%%%%%%%%%

\end{document}